\pdfoutput=1
\documentclass{bmvc2k}

\usepackage{multirow}
\usepackage{booktabs}
\usepackage{amsfonts}
\usepackage{colortbl}
\usepackage[ruled,vlined,linesnumbered]{algorithm2e}
\usepackage[figuresright]{rotating}

\usepackage{amssymb}

\usepackage{cleveref}

\title{PePESeg3D: Perception Prior Enhances Multi-Scale Segmentation for 3D Gaussian Splatting}

\addauthor{Sungjae Choi}{sungjae579@kaist.ac.kr}{1}
\addauthor{Seunghee Koh}{seunghee1215@kaist.ac.kr}{1}
\addauthor{Junmo Kim}{junmo.kim@kaist.ac.kr}{1}

\addinstitution{
 School of Electrical Engineering\\
 Korea Advanced Institute of Science and Technology (KAIST)\\
 Daejeon, South Korea
}

\runninghead{S. Choi et al.}{PePESeg3D}

\begin{document}

\maketitle

\begin{abstract}

Recent advancements in 3D Gaussian Splatting (3DGS) have extended its capabilities to multi-scale segmentation.
Existing methods reconstruct a scene with Gaussian primitives and learn multi-scale segmentation features separately, which leaves the geometry unaware of semantic structure and the feature learning dependent on incomplete mask supervision.
To address these limitations, we present \textbf{PePESeg3D}, a novel framework that injects perception priors into a multi-scale 3D Gaussian segmentation pipeline.
To fully exploit perception priors, we integrate them not only into contrastive feature learning but also into the upstream geometry reconstruction.
Specifically, PePE Reconstruction incorporates monocular depth and mask constraints to ensure semantically coherent object structures. Building on this aligned geometry, PePE Contrastive Learning leverages dense depth-color cues and view-consistent centroid supervision to compensate for the incompleteness of multi-scale masks obtained from a 2D foundation model.
Extensive experiments on the SPIn-NeRF, LERF-Mask, and NVOS benchmarks demonstrate that PePESeg3D achieves state-of-the-art performance in both multi-scale segmentation and scene reconstruction, highlighting the importance of integrating perception priors into both geometry optimization and feature learning for accurate multi-scale 3D segmentation.
Our code is available at \url{https://github.com/BeCow5X5/PePESeg3D}.
\end{abstract}
\section{Introduction}
The emergence of 3D Gaussian Splatting (3DGS) introduces a new paradigm in 3D scene understanding.
Unlike implicit neural representations, 3DGS explicitly represents scenes with discrete primitives.
By lifting the capabilities of 2D foundation models onto Gaussian primitives, 3DGS broadens its applicability not only for novel view synthesis but also for a wide range of tasks, such as semantic understanding~\cite{guo2024semantic}, and scene editing~\cite{Chen_2024_CVPR}.
The discrete nature of 3DGS facilitates segmentation~\cite{gaussian_grouping,zhu2025rethinking} by allowing specific semantic feature to be assigned to each Gaussian. This design naturally extends from single-scale segmentation to multi-scale understanding, where the compositions of primitives can represent varying granularities from fine parts to whole objects~\cite{ying2024omniseg3d,cen2025segment}. 

Typical single-scale segmentation methods jointly optimize Gaussian parameters and segmentation features, utilizing a unique ground truth segmentation map obtained from SAM~\cite{kirillov2023segany}. 
In contrast, multi-scale segmentation in 3DGS optimizes geometry and segmentation features sequentially, because joint optimization becomes unstable under complex ground truth maps with mask-specific granularity.
This design exposes two limitations.
First, primitives initialized solely via photometric supervision lack semantic awareness, leading to a geometric misalignment between Gaussian boundaries and object boundaries. 
Since segmentation features are subsequently learned on top of this fixed geometry, the misalignment propagates into the feature field and limits segmentation accuracy.
Second, the reliance on SAM limits the quality of multi-scale supervision.
Since SAM generates masks via grid-based point prompting, SAM often produces inconsistent mask granularity and omits entities depending on the view.
Multiple granularities further amplify these inconsistencies, leaving the contrastive learning stage with sparse and unreliable signals.


To address these limitations, we propose \textbf{PePESeg3D}, a multi-scale 3DGS segmentation framework that deeply integrates perception priors.
The framework is composed of two key components: Perception Prior Enhanced (PePE) Reconstruction and Perception Prior Enhanced (PePE) Contrastive Learning.
We first introduce PePE Reconstruction to resolve the spatial misalignment. 
It incorporates perception priors, 2D segmentation masks and monocular depth, into the geometry optimization.
Gaussian primitives are then initialized to minimize photometric error while respecting semantic boundaries and the underlying 3D geometry, which facilitates subsequent multi-scale feature learning.

Second, to group the geometrically refined 3D Gaussians into coherent multi-scale entities, we propose PePE Contrastive Learning.
We inject perceptual cues into the scale-aware mask supervision, which remedies the incompleteness of SAM masks.
Inspired by the observation that object boundaries are often discerned through abrupt changes in depth and color~\cite{liu2024sanerf}, we leverage these physical signals to supervise feature distinctiveness.
By modeling the distribution of depth and color differences, we provide dense supervision even in regions where SAM masks are ambiguous or missing.
Furthermore, we employ view-consistent centroid supervision that dynamically maintains global centroids at extreme scales to guide local features, encouraging label coherence across viewpoints.


Extensive experiments on real-world datasets demonstrate that PePESeg3D achieves state-of-the-art performance across multi-object, text-guided, and fine-grained segmentation tasks.
Notably, our framework demonstrates its versatility by not only outperforming existing multi-scale approaches but also remaining competitive with single-scale methods exclusively optimized for a single target granularity.
These results highlight the importance of perception-aware geometry optimization and feature learning for accurate and spatially coherent 3D segmentation.
In summary, our key contributions are as follows:

\begin{itemize}
    \item We propose \textbf{PePESeg3D}, a novel framework for multi-scale 3DGS segmentation that bridges the gap between geometry and semantics. 
    We introduce PePE Reconstruction, which integrates segmentation masks and depth priors to align Gaussian primitives with semantic boundaries.
    \item We propose PePE Contrastive Learning, a feature alignment mechanism that injects perceptual cues to refine feature distinctiveness.
    Complemented by view-consistent centroid supervision, it enables robust, physical scale-aware segmentation across varying viewpoints. 
    \item Our experiments empirically demonstrate that establishing a geometrically consistent foundation is essential for accurate multi-scale segmentation, by achieving state-of-the-art performance on the SPIn-NeRF and LERF-Mask benchmarks.
\end{itemize}

\section{Related Works}

\subsection{Single-Scale Segmentation in Radiance Fields}
Recent advances in 3D scene representation, driven by NeRF~\cite{mildenhall2020nerf} and 3DGS~\cite{kerbl3Dgaussians}, facilitate 3D segmentation~\cite{kobayashi2022distilledfeaturefields,tschernezki22neural,isrfgoel2023,cen2023segment,jsmbankILGS,shen2025trace3d,zhu2025objectgs}.
LERF~\cite{lerf2023} utilizes CLIP~\cite{radford2021learning} embedding distillation for language-driven region retrieval.
NVOS~\cite{ren-cvpr2022-nvos} leverages voxel feature embedding and MLP for user prompt segmentation.
Panoptic Lifting~\cite{Siddiqui_2023_CVPR} lifts 2D panoptic annotations to 3D by incorporating MLPs for panoptic segmentation.
Based on the faster optimization with explicit primitives, 3DGS emerges as a valuable representation for downstream tasks~\cite{xu2024depthsplat,wang2024gscream,flashsplat,jain2024gaussiancut,li2024gradiseggradientguidedgaussiansegmentation, ccllgs, opensplat3d} in 3D space.
Feature-3DGS~\cite{zhou2024feature} distills the encoder features of SAM~\cite{kirillov2023segany} into each Gaussian and uses the decoder for segmentation.
Gaussian Grouping~\cite{gaussian_grouping} employs supervision from a dense segmentation map generated by the tracking module~\cite{cheng2023tracking}. 
Unified-Lift~\cite{zhu2025rethinking} eliminates the need for mask association and clustering post-processing by learning instance identities in a single end-to-end training framework.

\subsection{Multi-Scale Segmentation in Radiance Fields}

Multi-scale segmentation methods aim to address the ambiguity where a pixel may belong to multiple semantic entities at different scales. 
One strategy is to utilize a finite set of discrete granularity levels provided by SAM. 
These scale levels are limited to predefined stages based on conceptual categories or hierarchical relationships, without modeling the physical size variation in 3D space~\cite{yang2026binary}.
OmniSeg3D~\cite{ying2024omniseg3d} introduces a discrete hierarchy of up to four levels and utilizes contrastive learning to align SAM masks into NeRF.
Click-Gaussian~\cite{choi2025click} presents a conceptual bi-level (fine, coarse) granularity feature field in 3DGS, trained via contrastive learning.
In contrast, other approaches represent scale as a continuous physical quantity, normalized to a range between 0 and 1, embedded in 3D geometry. 
This formulation allows the model to continuously select or interpolate between scales, enabling finer control over geometric consistency. 
GARField~\cite{garfield2024} leverages this by conditioning features on a continuous scale without explicitly separating levels. 
SAGA~\cite{cen2025segment} extends this to 3DGS using scale-gated affinity features, where a scale gate modulates the rendered features over continuous physical mask scales.
Building upon this continuous formulation, PePESeg3D leverages perception priors to enforce structural consistency, ensuring robust segmentation across both varying views and granularities.
\section{Preliminaries}
\subsection{3D Gaussian Splatting}
3D Gaussian Splatting (3DGS) represents a scene as a set $\mathcal{G}$ of 3D Gaussian primitives $g$, defined as: $g(\mathbf{x}) = \exp\{{-\frac{1}{2}(\mathbf{x}-\boldsymbol{\mu})^\top\boldsymbol{\Sigma}^{-1}(\mathbf{x}-\boldsymbol{\mu})}\}$. 
Each Gaussian is parametrized by a trainable mean $\boldsymbol{\mu}\in \mathbb{R}^3$ and a covariance $\boldsymbol{\Sigma}=\mathbf{R} \mathbf{S} \mathbf{S}^{\top} \mathbf{R}^{\top} \in \mathbb{R}^{3\times3}$, constructed from the scale $\mathbf{S} \in \mathbb{R}^3$ and the rotation matrix $\mathbf{R} \in \mathbb{R}^{3\times3}$.
Additional trainable parameters are opacity $\boldsymbol{\sigma} \in \mathbb{R}$, and color $\mathbf{c} \in \mathbb{R}^{48}$ represented by spherical harmonics~\cite{yu_and_fridovichkeil2021plenoxels}.
Given a random view image $I$ from a set of multi-view images, 3DGS projects $\mathcal{G}$ onto the 2D image plane.
The color $\mathbf{C}(p)$ for the pixel $p$ is computed by alpha blending \cite{KPLD21,kopanas2022neural} over $N_p$ depth-sorted Gaussians overlapped on $p$ following: $\mathbf{C}(p)=\sum_{i\in N_p} \mathbf{c}_i \alpha_i\prod_{j=1}^{i-1} (1 - \alpha_j)$, where $\alpha_i$ is the influence of the $i$-th Gaussian computed based on its opacity, 2D covariance, and distance to the pixel ~\cite{Yifan:DSS:2019}. All parameters of Gaussians are optimized by photometric loss~\cite{kerbl3Dgaussians} between $I$ and the rendered image.

\subsection{Gaussian Refinement}
\label{preliminary:gaussian_refinement}
A Gaussian $g$ overlapping the boundary of an arbitrary binary mask $M$ can be refined to $g^*$, which resides fully within the mask~\cite{hu2024semantic}. Projecting a 3D Gaussian onto the image plane yields a 2D covariance $\boldsymbol{\Sigma}'$, whose largest eigenpair is $(\lambda', \mathbf{v}')$.
For a $3\sigma$ confidence ellipse, the endpoints on the 2D major axis are $p'_{\pm} = \boldsymbol{\mu}' \pm 3\sqrt{\lambda'}\mathbf{v}'$.
Assuming $p_{+}'$ lies inside the mask and $p_{-}'$ outside, let $q'$ be the intersection between the segment $[p_{+}', p_{-}']$ with the mask boundary.
Under the local affine approximation of projection~\cite{964490}, the inner-mask ratio in 3D is approximated by its 2D counterpart: $r \approx r' = \frac{\lVert q' - p_{+}' \rVert}{\lVert p_{-}' - p_{+}' \rVert}$.
Using the 3D major axis direction $\mathbf{v}$ and the scale along this axis $\mathbf{S}_{\mathbf{v}}$, we refine the Gaussian as
\begin{equation}
    \mathbf{S}_{\mathbf{v}}^{*} = r \, \mathbf{S}_{\mathbf{v}},
    \qquad
    \boldsymbol{\mu}^{*} = p_{+} - \frac{1}{2}\mathbf{S}_{\mathbf{v}}^{*} \mathbf{v},
\end{equation}
where $p_+$ corresponds to the 3D endpoints before projection.
We construct the refined Gaussian $g^{*}$ by replacing only the mean and scale, while keeping other attributes unchanged.

\subsection{Scale-Aware Feature}
\label{preliminary:scale_aware_feature}
GARField~\cite{garfield2024} first defines the physical scale of a segmentation mask $M$ based on the spatial distribution of its 3D points.
Given the 3D camera coordinates $(X, Y, Z)$ of pixels $p \in M$, the scale $s_M$ is calculated as:
\begin{equation}
    s_M = 2 \sqrt{\mathrm{std}(X)^2 + \mathrm{std}(Y)^2 + \mathrm{std}(Z)^2}.
    \label{eq:preliminary_scale_calculation}
\end{equation}
To handle varying scale magnitudes, we normalize the raw scale $s$ into the range $[0, 1]$ based on the scene's minimum and maximum scales.
To incorporate this scale information, SAGA~\cite{cen2025segment} introduces the scale-aware feature.
Let $\mathbf{f} \in \mathbb{R}^D$ be the feature of a 3D Gaussian. The scale-aware feature $\mathbf{f}^s$ is defined by modulating $\mathbf{f}$ with a shared scale gate $\psi(s): [0,1] \rightarrow [0,1]^D$ via the Hadamard product $\odot$:
\begin{equation}
    \mathbf{f}^s = \psi(s) \odot \mathbf{f}.
\end{equation}

\section{Methods}
\begin{figure}[t!]
  \includegraphics[width=1.0\textwidth]{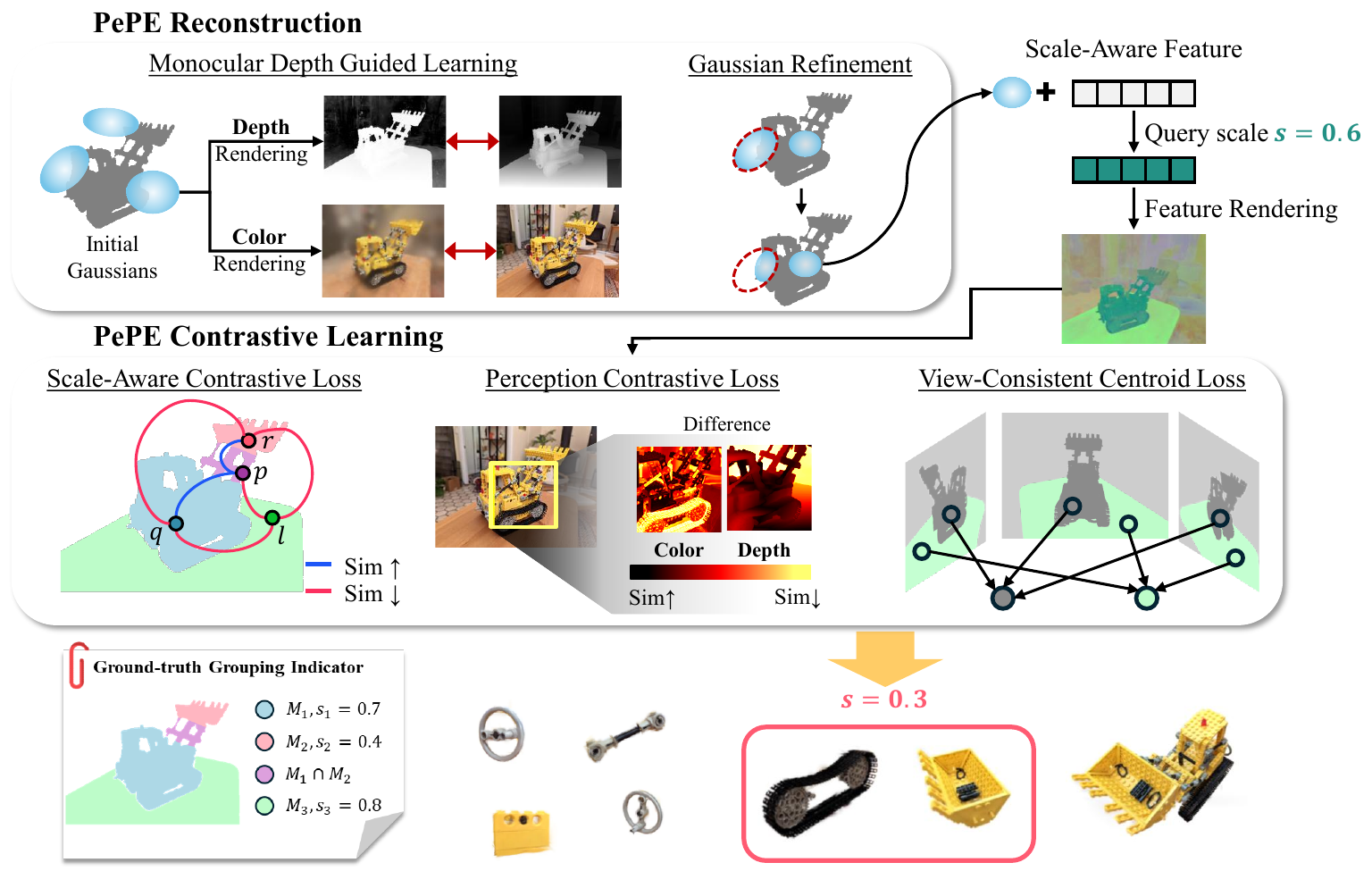}
  \caption{The two-stage pipeline of PePESeg3D.
  In PePE Reconstruction (top), we utilize monocular depth and segmentation masks to guide the Gaussian geometry, ensuring spatially coherent semantic structure. 
  In PePE Contrastive Learning stage (bottom), we optimize the feature field attached to the pretrained Gaussians using masks with assigned scale.
  By jointly using perceptual cues, scene-level centroids, and a grouping indicator that specifies hierarchical pixel associations, the framework enables precise, scale-conditioned 3D segmentation.
  }
  \label{fig:methods_architecture}
\end{figure}

Given a set of multi-view images, PePESeg3D reconstructs the scene into a 3D Gaussian representation that is aware of scale-dependent semantics and embedded with perception priors.
We extract a binary mask $M\in\{0,1\}^{H\times W}$ using SAM~\cite{kirillov2023segany} and a monocular depth map $D\in[0,1]^{H\times W}$ using Depth-Anything-V2~\cite{depth_anything_v2}, which serve as perception priors to guide our two-stage optimization pipeline, summarized in \Cref{fig:methods_architecture}.

The \emph{PePE Reconstruction} refines Gaussians to align with semantic boundaries and regularizes their depth distribution by monocular-depth guided learning, forming a geometric backbone. 
Next, we compute the physical scale of each mask and augment each Gaussian with additional features.
These features are optimized via \emph{PePE Contrastive Learning}, which combines the contrastive losses formulated using multi-scale masks, perceptual information, and view-consistent centroids.

\subsection{PePE Reconstruction}
\label{method:pepe_recon}
\subsubsection{Gaussian Refinement Guided Initialization}
We directly refine the Gaussian parameters to align with segment boundaries in the early stage of training.
For a given view with a set of masks $\mathcal{M}$, we first project 3D Gaussian $g$ with 3D covariance $\Sigma$ into a 2D Gaussian $g'$.
To effectively capture segments across varying scales, we organize the processing order based on the spatial extent of each mask.
We sort $\mathcal{M}$ in ascending order of the number of pixels in each mask, to prioritize fine-grained details.
We then iterate through the sorted mask sequence and determine whether $g'$ overlaps the boundary of $M_k \in \mathcal{M}$.
If the major axis $v'$ of $g'$ intersects the boundary of a mask $M_k$, we apply the Gaussian refinement described in \Cref{preliminary:gaussian_refinement}.
Once a Gaussian is refined, it is excluded from updates to prevent excessive splitting.

We also limit the refinement only to the Gaussians that lie close to surfaces so as not to disrupt the overall geometry, particularly for those in the background.
To identify such candidates, we measure the depth gap between the projected z coordinate $\mathbf{z}_k$ of the 3D Gaussian $g_k$ and the rendered depth of the pixel to which the mean of $g$ is projected.
The rendered depth is computed by alpha blending as $\hat{D}_p = \sum_{i \in N_p} \mathbf{z}_i \alpha_i\prod_{j=1}^{i-1} (1 - \alpha_j)$.
We then select a subset $\mathcal{G}_{R}$ consisting of the top 10\% of Gaussians with the smallest depth gaps, $| \hat{D}_p - \mathbf{z}_k |$.

\subsubsection{Monocular-Depth Constrained Learning}
As a geometric prior, we utilize monocular depth estimation~\cite{depth_anything_v2} to supervise the scene geometry.
Since monocular depth estimators provide relative depth without absolute scale, direct supervision is infeasible.
Following~\cite{midas}, we compute the optimal scale and shift $\alpha^*, \beta^*$ that minimize the least-squares error between the rendered depth map $\hat{D}$ and the monocular prior $D$.
The monocular depth loss is then defined as:
\begin{equation}
    \mathcal{L}_{depth} = \frac{1}{N_d}\sum_{p \in D}|\alpha^* \hat{D}_p + \beta^* - D_p|,
    \label{eq:method_depth_loss}
\end{equation}
where $N_d$ denotes the total number of pixels in the depth map.
This alignment ensures geometric consistency with the monocular prior and builds a geometrically coherent structure.

Combined with the photometric loss, the total objective is:
\begin{equation}
    \mathcal{L}_{recon} = (1-\lambda_{ph})\mathcal{L}_1+\lambda_{ph}\mathcal{L}_{\text{D-SSIM}} + \lambda_d\mathcal{L}_{depth},
    \label{eq:method_recon_loss}
\end{equation}
where $\lambda_{ph}$ and $\lambda_d$ are hyperparameters.
We optimize the 3DGS representation using this objective as a pretraining stage.
At iteration $K$, as primitives align with the depth map, we perform Gaussian refinement-guided initialization.
The resulting geometry then serves as a foundation for subsequent physical scale calculation and feature learning.

\subsection{PePE Contrastive Learning}
\label{method:pepe_contrastive}

In this stage, we optimize the feature $\mathbf{f}$ attached to the pretrained Gaussians and the scale gate $\psi$ to capture scale-dependent semantics by computing contrastive losses between rendered features.
For a pixel $p$ and a query scale $s$, the rendered feature $\mathbf{F}^s(p) = \sum_{i \in N_p} \mathbf{f}_i^s \alpha_i \prod_{j=1}^{i-1} (1 - \alpha_j)$ is obtained via alpha blending.
Given scale $s$, we utilize the cosine similarity $C^{s}_{pq}$ of their rendered features to quantify the semantic relationship between pixel pair $p$ and $q$:
\begin{equation}
    C^{s}_{pq} = \frac{\langle \mathbf{F}^s(p), \mathbf{F}^s(q) \rangle}{\| \mathbf{F}^s(p) \|_2 \| \mathbf{F}^s(q) \|_2}.
\end{equation}
We leverage $C^{s}_{pq}$ to pull semantically similar pairs together while pushing dissimilar ones apart.

\subsubsection{Scale-Aware Contrastive Loss}

We supervise the feature field using segmentation masks assigned according to the query scale, while preserving fine-grained correspondences.
We define a scale-dependent label function $\Lambda(s, p)$ that selects the most distinct object mask whose physical scale matches $s$.
Formally, let $\mathcal{M}_p = \{ M_k \in \mathcal{M} \mid p \in M_k, \ s_{M_k} \geq s \}$ be the set of masks at pixel $p$ that are larger than or equal to the query scale.
The active label is selected as a mask with the finest granularity within this set:
\begin{equation}
    \Lambda(s, p) = \arg\min\nolimits_{M_k \in \mathcal{M}_p} \{s_{M_k}\}.
\end{equation}
Based on this labeling, we construct the ground-truth grouping indicator $G^{s}_{pq}$.
We consider pixels $p$ and $q$ to belong to the same entity if they share the same label at scale $s$ or if they coexist in any mask finer than $s$.
Formally, utilizing the indicator function $\mathbf{1}[\cdot]$ which yields 1 if the enclosed condition holds and 0 otherwise, $G^{s}_{pq}$ is given by:
\begin{equation}
    G^{s}_{pq} = \mathbf{1}\Big[\left(\Lambda(s, p) = \Lambda(s, q)\right) \lor \left(\exists k : p, q \in M_k \land s_{M_k}<s \right) \Big].
\end{equation}
This formulation ensures that pixels belonging to the same small object are consistently treated as similar pairs, even when the query scale $s$ represents a larger entity, naturally enforcing hierarchical consistency.
The scale-aware contrastive objective is defined over the distribution of query scales $\mathcal{S}$ and the set of pixel pairs $\mathcal{P}$, where both the scale $s$ and the pixel pair $(p,q)$ are uniformly sampled from each domain:
\begin{equation}
    \label{eq:method_scale_loss}
    \mathcal{L}_{m}
    =\mathop{\mathbb{E}}\limits_{s\sim\mathcal{S}}\left[\mathop{\mathbb{E}}\limits_{(p,q)\sim \mathcal{P}}
    \Bigl[
    -\,G^{s}_{pq}\,C^{s}_{pq}
    +\bigl(1-G^{s}_{pq}\bigr)\,\max(C^{s}_{pq}, 0)
    \Bigr]\right].
\end{equation}

\subsubsection{Perception Contrastive Loss}

We formulate the objective that adjusts feature similarity based on photometric and geometric priors.
For each sampled scale $s \in \mathcal{S}$, we use a scale-adaptive window of radius $r(s)$ that increases with scale and consider pixel pairs $(p,q)$ satisfying both $\|p-q\|_2\le r(s)$ and $p,q \in I$.
To determine robust similarity criteria, we compute local statistics within the $r(s)$-neighborhood and establish adaptive thresholds for depth and color differences:
\[
\theta_d(p,q,s)=\max\bigl(\mu_d(p,s)+\sigma_d(p,s),\ \mu_d(q,s)+\sigma_d(q,s)\bigr),
\]
\[
\theta_c(p,q,s)=\max\bigl(\mu_c(p,s)+\sigma_c(p,s),\ \mu_c(q,s)+\sigma_c(q,s)\bigr),
\]
where $\mu_d$ and $\sigma_d$ denote the mean and standard deviation of absolute gradients computed on the relative depth map $D$.
These statistics are calculated individually over the local windows centered at $p$ and $q$.
Analogously, $\mu_c$ and $\sigma_c$ correspond to the statistics derived from the image $I$.

We then partition pixel pairs within the spatial window into the set of similar pairs $\mathcal{P}^{s}_{sim}$ and dissimilar pairs $\mathcal{P}^{s}_{dis}$.
A pair $(p, q)$ belongs to $\mathcal{P}^{s}_{sim}$ if both the depth difference and color difference fall below the adaptive thresholds $\theta_d(p,q,s)$ and $\theta_c(p,q,s)$, while it belongs to $\mathcal{P}^{s}_{dis}$ if both differences exceed their respective thresholds.
The perception contrastive loss pulls depth–color similar pairs together and pushes dissimilar pairs apart:
\begin{equation}
    \label{eq:method_perception_loss}
    \mathcal{L}_{\mathrm{p}}
    =\mathop{\mathbb{E}}\limits_{s \sim \mathcal{S}}\!\left[
    \mathop{\mathbb{E}}\limits_{(p,q)\sim\mathcal{P}^{s}_{\mathrm{dis}}}
    \!\bigl[\max(C^{s}_{pq}, 0)\bigr]
    -\mathop{\mathbb{E}}\limits_{(p,q)\sim \mathcal{P}^{s}_{\mathrm{sim}}}\!\bigl[C^{s}_{pq}\bigr]
    \right].
\end{equation}
To prioritize explicit mask supervision and prevent conflicting optimization objectives, we exclude pixel pairs from the perception loss computation if both pixels fall within valid mask regions.
The perception contrastive loss is exclusively applied to pixel pairs where neither pixel is masked or only one pixel is covered by a mask.
This objective augments mask supervision with perception priors without interfering with reliable mask supervision, while adapting spatial context to the physical scale.

\subsubsection{View-Consistent Centroid Loss}
To promote scene-level consistency beyond local, view-wise supervision, we construct two centroid sets at the extreme scales, an uppermost set $\mathcal{C}_U$ ($s=1.0$) and a lowermost set $\mathcal{C}_L$ ($s=0.01$).
From the uppermost scale gated feature maps $\mathbf{F}^{U}$, we extract candidate cluster means via HDBSCAN~\cite{McInnes2017} and merge candidates whose cosine similarity exceeds 0.9 using EMA updates to establish the final centroids for $\mathcal{C}_U$.
Analogously, we extract the lowermost centroids for $\mathcal{C}_L$ from $\mathbf{F}^{L}$.
We refresh these centroid sets every \(T\) iterations using random training views to guarantee exposure to global scale extremes.
For notation brevity, we use $\mathcal{C}^s(p, c)$ to denote the cosine similarity between the feature at pixel $p$ given scale $s$ and a centroid $c$.
For each extreme scale $s \in \{U, L\}$, we identify the nearest centroid $c_*^{s} = \operatorname*{arg\,max}_{c \in \mathcal{C}_s} \mathcal{C}^s(p, c)$.
Since centroids from random views may not match the current local view, we retain only pixels with sufficient similarity as $\mathcal{V}_s = \{ p \mid \mathcal{C}^s(p, c_*^{s}) > 0.9 \}$ to prevent erroneous attraction to unrelated prototypes.
The centroid loss is then formulated as:
\begin{equation}
\mathcal{L}_{\mathrm{c}} = \sum_{s \in \{U, L\}} \lambda_c^s \cdot \mathop{\mathbb{E}}\limits_{p \sim \mathcal{V}_s}
\Bigl[ - \mathcal{C}^s(p, c_*^{s}) \Bigr],
\end{equation}
where $\lambda_c^U$ and $\lambda_c^L$ balance the contribution of each extreme scale.

\subsubsection{Final Objective}
During the feature aggregation, a few unconstrained 3D features with excessively large norms can dominate the rendering, biasing the representation regardless of spatial relevance. 
To prevent this and ensure stable contrastive learning, it is crucial that all features reside on a unit hypersphere. 
Therefore, we regularize the norms of both the 3D Gaussian features and the aggregated 2D features toward unit length:
\begin{equation}
    \mathcal{L}_{\mathrm{r}} = \mathop{\mathbb{E}}\limits_{i \sim \mathcal{G}} \Bigl[ (\|\mathbf{f}_i\|_2 - 1)^2 \Bigr] + \mathop{\mathbb{E}}\limits_{p \sim I} \Bigl[ (\|\mathbf{F}(p)\|_2 - 1)^2 \Bigr],
\end{equation}
where $\mathcal{G}$ denotes the set of all 3D Gaussians and $I$ denotes the image pixels.

The final training objective is a weighted combination of proposed losses.
We define the total loss $\mathcal{L}_{\text{cont}}$ as follows:
\begin{equation}
    \mathcal{L}_{cont} = \mathcal{L}_m + \lambda_p \mathcal{L}_p + \mathcal{L}_c + \mathcal{L}_r,
    \label{eq:method_final_contrastive}
\end{equation}
where $\lambda_p$ is a hyperparameter.

\begin{figure}[t!]
    \centering
    \includegraphics[width=\linewidth]{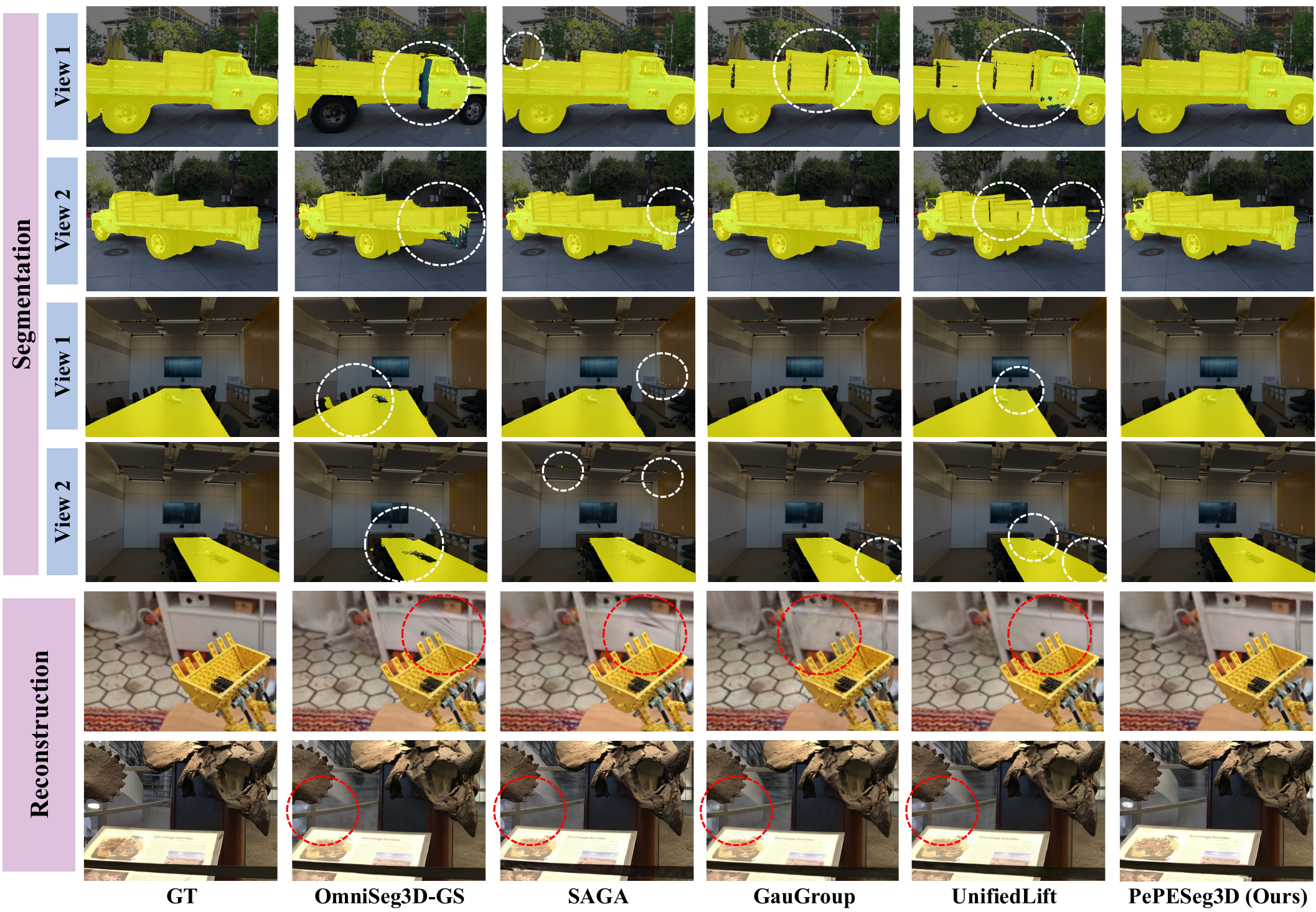}
    \vspace{-1.5em}
    \caption{Qualitative comparison of segmentation and reconstruction quality on SPIn-NeRF. The white circles indicate erroneous segmentation masks, while the red circles highlight geometric artifacts and blurred details in baseline segmentation methods.
    }
    \label{fig:exp_spin_nerf_joint}
    \vspace{-1.0em}
\end{figure}

\section{Experiments}
\subsection{Experimental Settings}
\subsubsection{Implementation Details}
In the PePE Reconstruction stage, each scene is trained for 30,000 iterations, with Gaussian refinement starting at the 4,000th iteration and lasting for twice the number of input views. 
We set $( \lambda_{ph}, \lambda_d )=(0.2, 0.05)$ and follow the original 3DGS for the remaining hyperparameters and optimization schedule.
In the PePE Contrastive Learning stage, geometry is frozen. 32-dimensional Gaussian features and a two-layer sigmoid-activated scale gate are optimized using Adam with a learning rate of 0.0025. 
The feature field is trained for 10,000 iterations using a batch of 1,000 sampled pixels and $(\lambda_p, \lambda_c^U, \lambda_c^L ) = (0.2, 0.3, 0.1)$. 
To stabilize the training process, we activate $\lambda_c^U$ and $\lambda_c^L$ starting from the 7,000th iteration and refresh centroid sets every 200 steps. 
We utilize SAM ViT-H~\cite{kirillov2023segany} for mask extraction and Depth-Anything-V2 ViT-B~\cite{depth_anything_v2} for depth estimation, and conduct all experiments on a single NVIDIA RTX 4090.

\subsubsection{Datasets} 
We evaluate on real-world datasets including SPIn-NeRF~\cite{spinnerf}, NVOS~\cite{ren-cvpr2022-nvos}, LERF-Mask~\cite{gaussian_grouping}, and LERF-Mask-Fine~\cite{choi2025click}.
These datasets collectively span a range of annotation granularities.
SPIn-NeRF provides ground-truth object masks that include both whole objects and fine-grained subparts across most viewpoints.
We construct the test split by holding out every eighth image to evaluate reconstruction on novel views and segmentation across annotated views.
NVOS provides detailed ground-truth masks for two views from the LLFF~\cite{mildenhall2019llff} dataset, enabling the evaluation of performance on intricate details.
For NVOS, we adopt the identical split used for SPIn-NeRF to evaluate novel view synthesis and train the segmentation model.
LERF-Mask provides text-paired masks corresponding to coarse, object-level entities, while LERF-Mask-Fine targets fine-grained parts within objects.
For both, we follow the official splits, using 150--250 views for training and 3--4 views for testing to assess generalization to unseen views.

\begin{figure}[t!]
    \includegraphics[width=\linewidth]{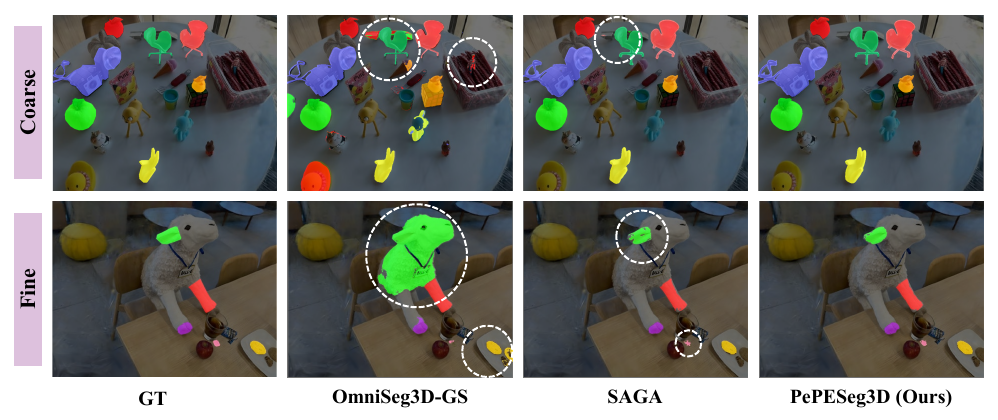}
    \vspace{-2.0em}
    \caption{Qualitative comparison on LERF-Mask and LERF-Mask-Fine datasets. The white circles indicate regions where baseline methods incorrectly merge distinct objects or split a single entity.
    }
    \label{fig:exp_lerf_mask_multiscale}
\end{figure}

\begin{table}
    \begin{center}
    \resizebox{\linewidth}{!}{%
    \begin{tabular}{l c cc ccc c} 
        \toprule
        \multirow{2}{*}{\textbf{Model}} 
        & \multirow{2}{*}{\textbf{Type}} 
        & \multicolumn{2}{c}{Segmentation} 
        & \multicolumn{3}{c}{Reconstruction}
        & \multirow{2}{*}{\textbf{Training Time}} \\
        
        \cmidrule(lr){3-4}\cmidrule(lr){5-7}
        
            & 
            & \textbf{mIoU}$\uparrow$ (\%)
            & \textbf{mAcc}$\uparrow$ (\%)
            & \textbf{PSNR}$\uparrow$ (dB)
            & \textbf{SSIM}$\uparrow$ 
            & \textbf{LPIPS}$\downarrow$ 
            & \\ 
            \midrule
        
        GauGroup~\cite{gaussian_grouping} & Single & 84.9 & 97.5 & 26.46 & 0.846 & 0.154 & 32 min \\
        
        UnifiedLift~\cite{zhu2025rethinking} & Single & 85.0   & \underline{97.8} & \underline{26.63} & \underline{0.854} & \underline{0.146} & 72 min \\
        
        \midrule
        
        OmniSeg3D-GS~\cite{ying2024omniseg3d} & Multi & 82.5 & 97.4 & 26.05 & 0.850 & 0.153 & 44 min \\
                                          
        SAGA~\cite{cen2025segment} & Multi & \underline{91.9} & \textbf{98.9} & 26.09 & 0.850 & 0.153 &  23 min \\
        
        \textbf{PePESeg3D (Ours)} & Multi  & \textbf{92.3} & \textbf{98.9} & \textbf{26.81} & \textbf{0.857} & \textbf{0.142} & 32 min \\
        \bottomrule
    \end{tabular}
    }
    \end{center}
    \caption{Joint segmentation and reconstruction results on SPIn-NeRF. \textbf{Best} and \underline{second-best} results are highlighted. PePESeg3D achieves state-of-the-art performance across both segmentation and reconstruction metrics.}
    \label{tab:exp_spin_joint_results}
\end{table}

\begin{table*}[t!]
    \setlength{\tabcolsep}{1pt}
    \begin{minipage}[t]{0.66\linewidth}
    \centering
        \begin{minipage}[t]{0.48\linewidth}
            \centering
            \resizebox{\linewidth}{!}{%
                \begin{tabular}{l cc}
                    \toprule
                    \textbf{Method} & \textbf{mIoU}$\uparrow$ & \textbf{mBIoU}$\uparrow$ \\
                    \midrule
                    Panoptic-Lifting-GS~\cite{Siddiqui_2023_CVPR} & 70.7 & 65.8 \\
                    GauGroup~\cite{gaussian_grouping} & 72.8 & 67.6 \\
                    CCL-LGS~\cite{ccllgs} & 72.5 & 67.8 \\
                    Gaga~\cite{lyu2024gaga} & 74.7 & 72.2 \\
                    UnifiedLift~\cite{zhu2025rethinking} & \underline{80.9} & \underline{77.1} \\
                    OpenSplat3D~\cite{opensplat3d} & \textbf{83.5} & \textbf{78.4} \\
                    \bottomrule
                \end{tabular}%
            }\\[3pt]
            (a) Single-Scale
        \end{minipage}%
        \hfill
        \begin{minipage}[t]{0.48\linewidth}
            \centering
            \resizebox{\linewidth}{!}{%
                \begin{tabular}{l cc}
                    \toprule
                    \textbf{Method} & \textbf{mIoU}$\uparrow$ & \textbf{mBIoU}$\uparrow$ \\
                    \midrule
                    OmniSeg3D-GS~\cite{ying2024omniseg3d} & 74.7 & 71.8 \\
                    SAGA~\cite{cen2025segment} & 78.4 & 74.0 \\
                    \textbf{PePESeg3D (Ours)} & \textbf{80.5} & \textbf{76.5} \\
                    \bottomrule
                \end{tabular}%
            }\\[3pt]
            (b) Multi-Scale
        \end{minipage}
        \vspace{10pt}
        \caption{Segmentation performance on LERF-Mask across single-scale and multi-scale methods.}
        \label{tab:exp_lerf_mask_segmentation_results}
    \end{minipage}
    \hfill
    \begin{minipage}[t]{0.32\linewidth}
        \centering
        \setlength{\tabcolsep}{1pt}
        \resizebox{\linewidth}{!}{%
            \begin{tabular}{l cc}
                \toprule
                \textbf{Method} & \textbf{mIoU}$\uparrow$ & \textbf{mBIoU}$\uparrow$ \\
                \midrule
                OpenSplat3D~\cite{opensplat3d} & 15.0 & 14.6 \\
                UnifiedLift~\cite{zhu2025rethinking} & 16.0 & 14.7 \\
                CCL-LGS~\cite{ccllgs} & 23.1 & 22.1 \\
                OmniSeg3D-GS~\cite{ying2024omniseg3d} & 39.8 & 37.3 \\
                SAGA~\cite{cen2025segment} & \underline{69.6} & \underline{67.8} \\
                \textbf{PePESeg3D (Ours)} & \textbf{70.8} & \textbf{68.3} \\
                \bottomrule
            \end{tabular}%
        }
        \vspace{10pt}
        \caption{Segmentation results on LERF-Mask-Fine.}
        \label{tab:exp_lerf_mask_fine_segmentation_results}
    \end{minipage}
\end{table*}

\subsubsection{Evaluation Protocols}
For SPIn-NeRF and LERF-Mask, we follow the label propagation protocol~\cite{gaussian_grouping} and report IoU, pixel accuracy (Acc), and Boundary IoU (BIoU).
We extract a reference centroid of the annotated object using HDBSCAN and identify corresponding centroids in other views whose feature similarity with reference centroids exceeds 0.7.
For NVOS and LERF-Mask-Fine, we adopt a promptable segmentation approach by averaging points sampled from the target mask in a reference view and identifying the target object across views by retrieving pixels with feature similarity exceeding 0.9.
Reconstruction quality is evaluated using PSNR, SSIM~\cite{1284395}, and LPIPS~\cite{zhang2018unreasonable}.
All baselines are reproduced and evaluated using their official implementation and reported evaluation protocols for fair comparison.

\begin{figure}[t!]
\begin{center}
    \includegraphics[width=0.95\linewidth]{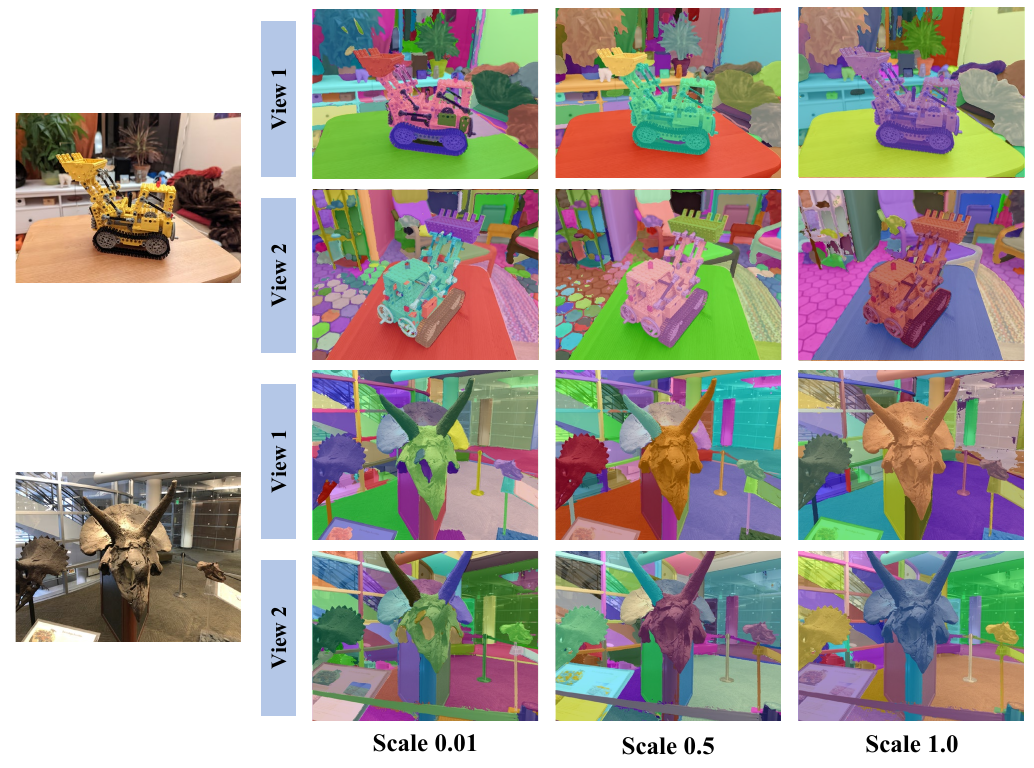}
    \vspace{-1.0em}
    \caption{Qualitative visualization of multi-scale semantic clusters extracted by PePESeg3D. Distinct colors represent unique clusters identified via HDBSCAN. The clustering demonstrates accurate scale-adaptive granularity and robust view consistency.}
    \label{fig:automatic}
\end{center}
\vspace{-1.0em}
\end{figure}

\subsection{Segmentation Results}

As presented in \Cref{tab:exp_spin_joint_results}, PePESeg3D demonstrates the best performance on the SPIn-NeRF dataset by achieving mIoU of 92.3\% and mAcc of 98.9\%.
Particularly, the 7.3 percent point (\%p) mIoU gap over the leading single-scale method, UnifiedLift, reflects a structural limitation of single-scale approaches to capture fine-grained objects.
Qualitatively, as highlighted by the white circles in \Cref{fig:exp_spin_nerf_joint}, each baseline exhibits distinct failure modes.
SAGA merges the truck and desk with semantically distinct backgrounds, illustrating the difficulty of resolving object boundaries in multi-scale feature learning.
Gaussian Grouping and UnifiedLift tend to produce fragmented masks that under-segment a single object, reflecting the limited expressiveness of their coarse supervision.
In contrast, PePESeg3D groups each target as a coherent entity, which we attribute to the semantic-aware reconstruction, combined with a robust understanding of multi-granularity semantics and view consistency.

The LERF-Mask dataset results in \Cref{tab:exp_lerf_mask_segmentation_results} demonstrate that PePESeg3D establishes state-of-the-art performance among multi-scale methods, surpassing the previous best method SAGA by margins of 2.1\%p in mIoU and 2.5\%p in mBIoU.
Unlike SPIn-NeRF, where mixed-granularity annotations favor multi-scale approaches, LERF-Mask evaluates on coarse targets that naturally suit single-scale pipelines.
PePESeg3D remains competitive with recent single-scale methods such as UnifiedLift and OpenSplat3D, which are designed exclusively for certain granularity.
These results show that our method effectively operates across multiple scales without sacrificing accuracy at any particular granularity.
This performance advantage extends to the LERF-Mask-Fine dataset reported in \Cref{tab:exp_lerf_mask_fine_segmentation_results}, where annotations target only fine-grained parts.
Single-scale methods optimize a fixed object-level target, so their accuracy collapses to 15--23\% mIoU.
Our method attains 70.8\% mIoU, outperforming SAGA by 1.2\%p.
Qualitatively, as visualized in \Cref{fig:exp_lerf_mask_multiscale}, PePESeg3D produces clean segmentation boundaries for coarse objects while preserving fine-grained semantic details for small parts.
Finally, we evaluate our method on the NVOS dataset to verify its effectiveness in capturing highly intricate details.
As summarized in \Cref{tab:nvos_segmentation_results}, PePESeg3D outperforms both traditional NeRF-based models and recent 3DGS models, achieving the highest mIoU of 92.2\% and mAcc of 98.5\%.

\begin{table*}[t!]
    \setlength{\tabcolsep}{4pt}
    \renewcommand{\arraystretch}{1.05}
    \centering
    \begin{minipage}[t]{0.45\linewidth}
        \centering
        \resizebox{0.9\linewidth}{!}{%
            \begin{tabular}{lcc}
                \toprule
                \textbf{Method} & \textbf{mIoU (\%)} & \textbf{mAcc (\%)} \\
                \midrule
                NVOS~\cite{ren-cvpr2022-nvos} & 70.1 & 92.0 \\
                ISRF~\cite{isrfgoel2023} & 83.8 & 96.4 \\
                SA3D~\cite{cen2023segment} & 90.3 & 98.2 \\
                OmniSeg3D~\cite{ying2024omniseg3d} & \underline{91.7} & \underline{98.4} \\
                \bottomrule
            \end{tabular}%
        }\\[3pt]
        (a) NeRF model
    \end{minipage}%
    \begin{minipage}[t]{0.45\linewidth}
        \centering
        \resizebox{0.9\linewidth}{!}{%
            \begin{tabular}{lcc}
                \toprule
                \textbf{Method} & \textbf{mIoU (\%)} & \textbf{mAcc (\%)} \\
                \midrule
                GauGroup~\cite{gaussian_grouping} & 85.6 & 97.3 \\
                SAGA~\cite{cen2025segment} & 91.5 & 98.3 \\
                \textbf{PePESeg3D (Ours)} & \textbf{92.2} & \textbf{98.5} \\
                \bottomrule
            \end{tabular}%
        }\\[3pt]
        (b) 3DGS model
    \end{minipage}
    \vspace{10pt}
    \caption{Segmentation results on NVOS dataset. PePESeg3D outperforms both traditional NeRF-based approaches and recent 3DGS models on preserving intricate details.}
    \label{tab:nvos_segmentation_results}
\end{table*}

\begin{table*}[t!]
    \setlength{\tabcolsep}{6pt}
    \renewcommand{\arraystretch}{1.05}
    \centering
    \resizebox{0.5\linewidth}{!}{%
        \begin{tabular}{lccc}
            \toprule
            \textbf{Model} & \textbf{PSNR}$\uparrow$ (dB) & \textbf{SSIM}$\uparrow$ & \textbf{LPIPS}$\downarrow$ \\
            \midrule
            GauGroup~\cite{gaussian_grouping}     & \underline{26.71} & 0.839 & \underline{0.153} \\
            UnifiedLift~\cite{zhu2025rethinking}  & 26.48 & \underline{0.843} & \underline{0.153} \\
            OmniSeg3D-GS~\cite{ying2024omniseg3d} & 25.71 & 0.837 & 0.162 \\
            SAGA~\cite{cen2025segment}            & 25.86 & 0.837 & 0.161 \\
            \textbf{PePESeg3D (Ours)}             & \textbf{27.08} & \textbf{0.850} & \textbf{0.145} \\
            \bottomrule
        \end{tabular}%
    }
    \vspace{10pt}
    \caption{Reconstruction results on NVOS.}
    \label{tab:exp_llff_reconstruction_results}
\end{table*}

\subsection{Reconstruction Results}
\begin{figure}[t!]
  \begin{center} 

    \begin{minipage}{\linewidth}
        \centering
        \includegraphics[width=0.65\linewidth]{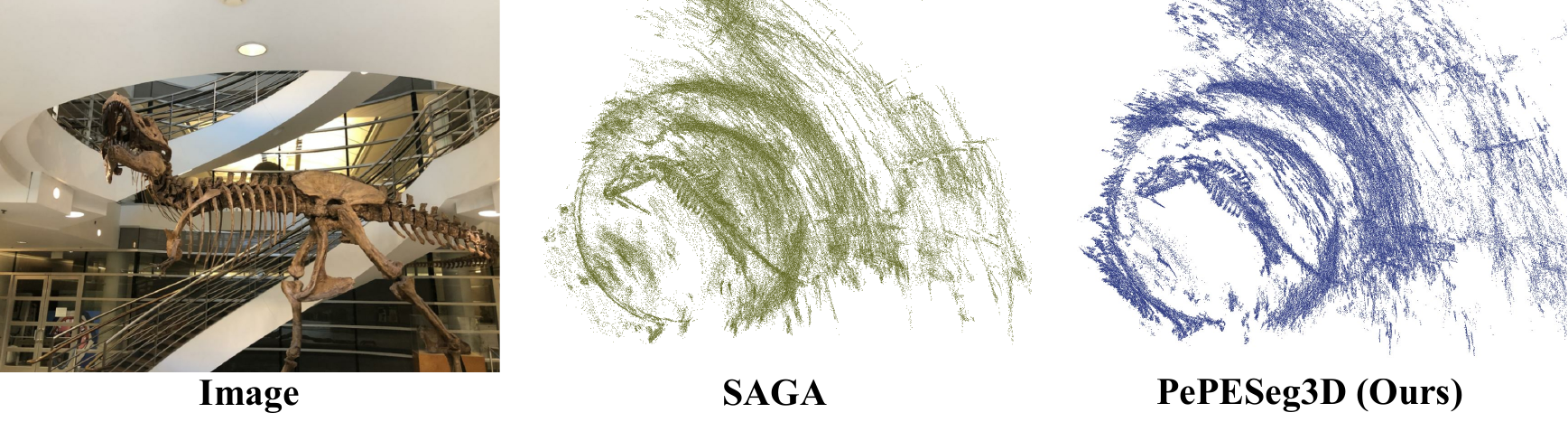} \\
        (a) Geometric noise mitigation (NVOS T-Rex)
    \end{minipage}
    \begin{minipage}{0.28\linewidth}
        \centering
        \includegraphics[width=\linewidth]{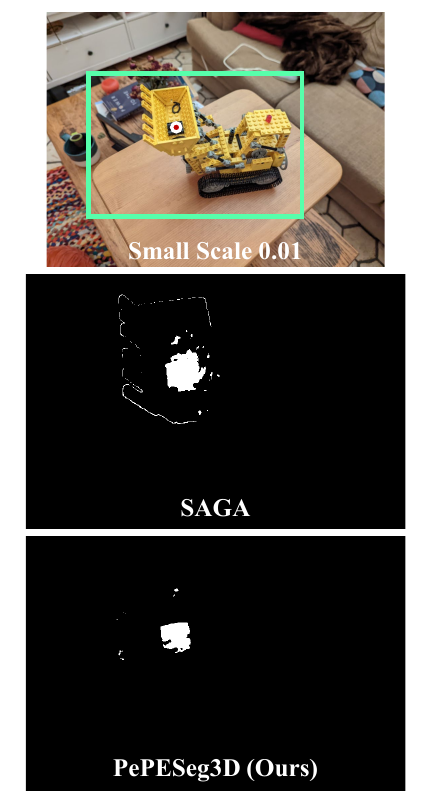} \\
        (b) Fine-grained segmentation (SPIn-NeRF Lego)
    \end{minipage}
    \hfill 
    \begin{minipage}{0.70\linewidth}
        \centering
        \includegraphics[width=\linewidth]{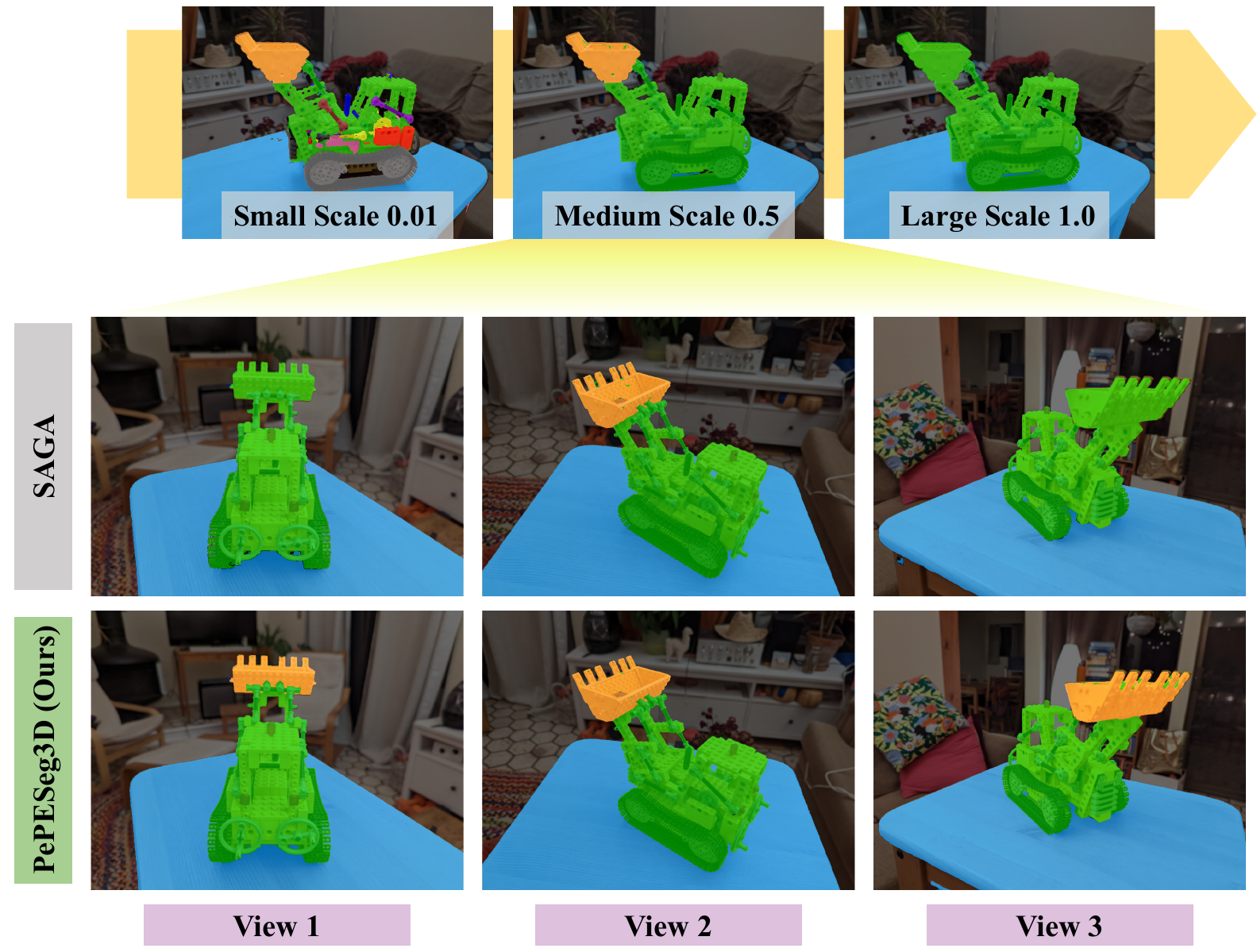} \\
        (c) Global scale consistency (SPIn-NeRF Lego)
    \end{minipage}

\end{center}
\caption{\textbf{Qualitative analysis of PePESeg3D.} 
(a) By incorporating depth constraints, our method effectively removes floating artifacts and geometric noise. 
(b) Our method accurately segments fine-grained details even at smaller object scales. 
(c) While the baseline exhibits view-dependent scale ambiguity, our method preserves strict global consistency.
}
\label{fig:qualitative_vertical}
\end{figure}

To verify whether the proposed components produce a semantically coherent foundation, we also evaluate reconstruction quality.
The quantitative results in \Cref{tab:exp_spin_joint_results,tab:exp_llff_reconstruction_results} demonstrate that PePESeg3D achieves superior reconstruction quality across datasets.
On SPIn-NeRF dataset, our method attains the highest performance across all metrics surpassing the previous best multi-scale method.
On NVOS dataset, it further improves upon SAGA by 1.22 dB.
More surprisingly, PePESeg3D surpasses even single-scale methods that explicitly inject segmentation information into scene geometry through joint optimization.
As highlighted by the red circled regions in \Cref{fig:exp_spin_nerf_joint}, baseline methods suffer from artifacts and blurred details.
In contrast, our method effectively suppresses these artifacts, demonstrating enhanced geometric alignment along object boundaries while successfully recovering details in background regions.
Furthermore, as shown in \Cref{fig:qualitative_vertical}-(a), incorporating monocular depth priors significantly improves overall geometry, eliminating the noisy Gaussians between the Trex skeleton and the background wall.
These qualitative results validate that our approach improves not only segmentation accuracy but also scene reconstruction fidelity.

\begin{table*}[t!]
    \begin{minipage}[t]{0.50\linewidth}
        \begin{center}
            \resizebox{\linewidth}{!}{
                \begin{tabular}{lccc}
                    \toprule
                    \textbf{Method} & \textbf{PSNR}($\uparrow$) & \textbf{SSIM}($\uparrow$) & \textbf{LPIPS}($\downarrow$) \\
                    \midrule
                    Baseline & 25.86 & 0.837 & 0.161 \\
                    + Gaussian Refinement & 26.32 & 0.843 & 0.154 \\
                    \rowcolor{gray!10} 
                    \textbf{+ Mono Depth (Full Model)} & \textbf{27.08} & \textbf{0.850} & \textbf{0.145} \\
                    \bottomrule
                \end{tabular}
            }
        \end{center}
        \caption{Ablation of PePE Reconstruction on NVOS.}
        \label{tab:exp_ablation_reconstruction}
    \end{minipage}
    \hfill 
    \begin{minipage}[t]{0.48\linewidth}
        \begin{center}
            \resizebox{\linewidth}{!}{
                \begin{tabular}{lcc}
                    \toprule
                    \textbf{Method} & \textbf{mIoU (\%)} & \textbf{mBIoU (\%)} \\
                    \midrule 
                    Baseline & 78.4 & 74.0 \\
                    + PePE Reconstruction          & 79.2 & 75.3 \\
                    + Perception Loss               & 80.2 & 76.0 \\
                    \rowcolor{gray!10} 
                    \textbf{+ Consistency (Full Model)} & \textbf{80.5} & \textbf{76.5} \\
                    \bottomrule
                \end{tabular}
            }
        \end{center}
        \caption{Ablation of PePESeg3D on LERF-Mask.}
        \label{tab:exp_ablation_segmentation}
    \end{minipage}
\vspace{-1.0em}
\end{table*}

\subsection{Analysis}

\paragraph{Ablation Studies}
We conduct ablation studies to validate the contribution of each component in PePESeg3D.
Regarding reconstruction results in \Cref{tab:exp_ablation_reconstruction}, incorporating Gaussian refinement leads to a performance gain of 0.46 dB in PSNR, as it precisely aligns primitives to capture fine details.
Subsequently, adding monocular depth supervision yields an additional increase of 0.76 dB by regularizing the scene geometry in regions where photometric supervision is insufficient.
These improvements demonstrate that both components effectively contribute to high-fidelity rendering.

In \Cref{tab:exp_ablation_segmentation}, we evaluate the contribution of each component to the final segmentation performance by utilizing the complete pipeline that unifies reconstruction and contrastive learning.
While all proposed components positively contribute to the performance improvement, we observe that the perception loss and PePE Reconstruction play the most significant roles.
The PePE Reconstruction boosts performance by an additional 0.8\%p, confirming that our geometric refinement establishes a robust structural foundation essential for multi-scale segmentation.
The perception loss yields a substantial improvement of 1.0\%p, demonstrating its effectiveness in facilitating feature learning through depth and color priors. 
Finally, the view-consistency loss provides a gain of 0.3\%p in mIoU, encouraging label consistency across different viewpoints.
These results confirm that the coupling of the geometric reconstruction stage and the perception-guided contrastive learning stage creates a powerful synergy, enabling highly precise and robust multi-scale segmentation.

\paragraph{Qualitative Analysis}
We further examine qualitative behavior across datasets in \Cref{fig:automatic} and \Cref{fig:qualitative_vertical}.
\Cref{fig:automatic} visualizes the robust multi-scale feature representations learned by PePESeg3D. 
As the query scale transitions from 1.0 (coarse) to 0.01 (fine), the semantic clusters smoothly adapt from encompassing whole objects to distinguishing individual parts.
Notably, consistent semantic labels are maintained across varying camera perspectives. 
As shown in \Cref{fig:qualitative_vertical}-(b), when querying target objects at minimal scale, SAGA struggles to isolate fine structures, yielding noisy masks.
In contrast, our method sharply extracts the exact target components.
Beyond local precision, \Cref{fig:qualitative_vertical}-(c) compares multi-scale segmentation across viewpoints.
SAGA assigns different segments to the same object depending on the view, while PePESeg3D maintains coherent segmentation across all three views and scales.
This visually confirms that our framework effectively disentangles spatial granularity while preserving 3D view consistency.

\section{Conclusion}

We present PePESeg3D, a multi-scale 3DGS segmentation framework that integrates perception priors into geometry reconstruction and contrastive feature learning.
By injecting perception priors into both stages, PePESeg3D ensures that semantic structure is established at the geometry level and leveraged in feature learning.
PePE Reconstruction uses monocular depth and segmentation masks to ground the scene geometry in semantic structure. 
Subsequently, PePE Contrastive Learning groups the primitives using dense depth-color cues and global centroid supervision to compensate for incomplete multi-scale SAM masks.
These components together yield strong empirical performance across diverse benchmarks.
PePESeg3D achieves state-of-the-art segmentation and reconstruction on SPIn-NeRF and NVOS.
On LERF-Mask, it surpasses all multi-scale methods while remaining competitive even with single-scale approaches. 
It further outperforms the previous multi-scale method on LERF-Mask-Fine, demonstrating robust performance across both coarse and fine granularities.
We further verify the effectiveness of our design through ablation studies, showing that the reconstruction and contrastive stages act synergistically.
The semantically aligned reconstruction improves segmentation and the perception contrastive loss provides the largest individual gain on accuracy.
These findings underscore that integrating perception priors into both geometry and feature learning stages is essential for accurate multi-scale 3DGS segmentation.

\section*{Acknowledgements}
This research was supported by Basic Science Research Program through the National Research Foundation of Korea(NRF) funded by the Ministry of Education(RS-2025-25404201).
This work was supported by the National Research Foundation of Korea(NRF) grant funded by the Korea government(MSIT) (RS-2026-25470670).

\bibliography{egbib}
\newpage
\appendix
\section*{Appendix for PePESeg3D: Perception Prior Enhances Multi-Scale Segmentation for 3D Gaussian Splatting}

\section{Additional Implementation Details for PePE Reconstruction}
To ensure stable optimization, we adapt the depth loss weight $\lambda_d$ based on the camera configuration of each scene.
In scenes captured with $360^\circ$ trajectories, the abundance of multi-view cues typically provides sufficient geometric constraints. 
Consequently, excessive reliance on monocular depth priors may conflict with the true geometry due to potential scale inconsistencies across wide viewpoints.
Therefore, we set a lower weight of $\lambda_d = 0.05$ for $360^\circ$ scenes to mitigate this interference. 
For forward-facing scenes where geometric ambiguities are more prevalent, we increase the weight to $\lambda_d = 0.1$ to leverage stronger geometric supervision from the monocular prior.
As training progresses, we employ a weight decay schedule, gradually reducing $\lambda_d$ to $0.02$ as training progresses.
This strategy relaxes the geometric constraints in later iterations, allowing the model to focus on minimizing photometric error and refining high-frequency textures.

While the monocular depth loss in the main text (Eq.~\ref{eq:method_depth_loss}) utilizes global alignment to address the scale ambiguity of relative depth maps, relying solely on a global context may overlook local geometric details. 
To provide more precise geometric supervision, we extend this formulation by incorporating a local alignment constraint.
Specifically, for each pixel, we define a local window $\Omega_w$ with radius $r_w$. 
Within this window, we compute the local optimal scale $\alpha_w^*$ and shift $\beta_w^*$ that minimize the least-squares error between the rendered depth $\hat{D}$ and the monocular prior $D$. 
This allows the supervision to adapt to local geometric inconsistencies that a global alignment might miss. 
The enhanced depth loss, combining both global and local terms, is defined as:
\begin{equation}
    \mathcal{L}_{depth} = \mathcal{L}_{global}
    + 
    \frac{1}{|\Omega_w|}\sum_{p \in \Omega_w} | \alpha_w^* \hat{D}_p + \beta_w^* - D_p |,
    \label{eq:method_depth_loss_with_local}
\end{equation}
where $\mathcal{L}_{global}$ corresponds to ~\Cref{eq:method_depth_loss}.

\section{Additional Implementation Details for PePE Contrastive Learning}

\subsection{Sampling Strategy}
While we primarily employ uniform sampling for query scales and pixel pairs, we introduce a hard example mining strategy to enhance discriminability in ambiguous regions.
We identify hard sample pairs as those that satisfy the grouping condition solely based on the active label at scale $s$, without being co-located in any finer-grained mask. Formally, a hard sample pair $(p, q)$ satisfies $\Lambda(s, p) = \Lambda(s, q)$ and does not belong to any common finer mask ($\nexists k : p, q \in M_k \land s_{M_k} < s $).
These pairs represent cases where the grouping changes according to the scale, making them more challenging to learn and crucial for capturing multi-scale semantics. 
To address this, we augment the standard uniform sampling by explicitly oversampling these hard pairs within each training batch.
Regarding the scale-aware contrastive loss (Eq.~\ref{eq:method_scale_loss}), we exclude pixels that are not covered by any mask instance from the sampling process. 
This constraint ensures that the model focuses on learning valid mask correspondences and prevents erroneous signals arising from ambiguous regions where no mask definitions exist.
Additionally, to encourage local spatial consistency, we average each Gaussian's feature with those of its 16 nearest neighbors before rendering, smoothing out isolated feature outliers in 3D space.

\subsection{Reweighting Strategy}

To appropriately balance the contribution of masks during contrastive learning, we adopt the reweighting strategy from \cite{cen2023segment}. 
Since pixels belonging to large masks are more frequently sampled, they tend to dominate the optimization process. 
To mitigate this, we define a pixel-wise weight $\omega(p)$ as the inverse of the average area of masks containing pixel $p$:
\begin{equation}
    \omega(p) = \Bigl( \frac{1}{|\mathcal{M}_p|} \sum_{M \in \mathcal{M}_p} |M| \Bigr)^{-1},
    \label{eq:appendix_mask_reweighting}
\end{equation}
where $\mathcal{M}_p = \{ M \in \mathcal{M} \mid M(p) = 1 \}$ denotes the set of masks covering pixel $p$, and $|M|$ represents the mask area defined as the number of pixels. 
For a sampled pixel pair $(p, q)$, the final balancing weight is given by $\omega(p) \cdot \omega(q)$.
To stabilize training, all weights are min-max normalized to the range $[1, 10]$ in each training iteration.
This weighting scheme ensures that pixel pairs from small masks are not underrepresented during contrastive optimization. 
By applying this weighting term, the scale-aware contrastive loss (Eq.~\ref{eq:method_scale_loss}) is formally defined as:

\begin{equation}
\label{eq:scale_feat}
\mathcal{L}_{m}
=\mathop{\mathbb{E}}\limits_{s\sim\mathcal{S}}\left[\mathop{\mathbb{E}}\limits_{(p,q)\sim \mathcal{P}}
w_{pq}\Bigl(
-\,G^{s}_{pq}\,C^{s}_{pq}
+\bigl(1-G^{s}_{pq}\bigr)\,\max(C^{s}_{pq}, 0)
\Bigr)\right].
\end{equation}

\begin{figure}[t]
  \centering
    \includegraphics[width=0.6\linewidth]{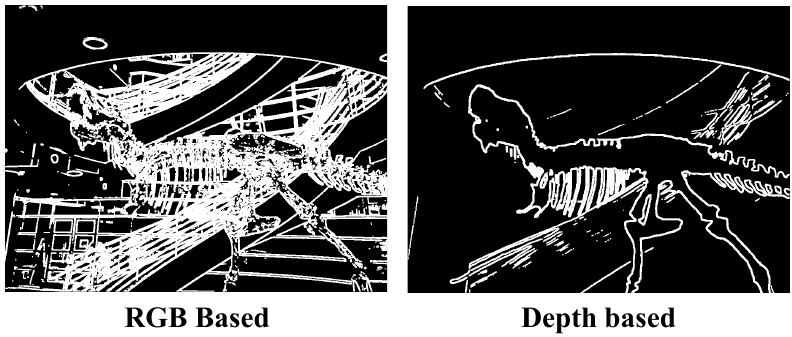}
  \caption{Comparison on Boundary Map Formulation.}
  \label{fig:appendix_boundary_map}
\end{figure}

\subsection{Implementation Details in Perception Loss}

While perception priors provide dense semantic signal, they often contain noise, particularly near object boundaries where feature transitions are ambiguous. 
To mitigate this, we restrict the loss computation to valid interior regions, excluding noisy boundaries.
For robust boundary detection, we utilize the monocular depth map $D$ rather than the RGB image. 
As illustrated in~\Cref{fig:appendix_boundary_map}, gradient computation on RGB images tends to excessively capture fine textures and subtle lighting variations, resulting in false boundaries even within homogeneous regions of a single object. 
In contrast, depth maps provide a more abstracted, geometry-aligned representation of the object structure.
Therefore, we construct the boundary map using a depth-based approach. 
Using relative depth map $D$, we form a boundary map by thresholding the gradient~\cite{996} magnitude and dilating. The interior region is then defined as its complement.

\begin{algorithm}[!h]
\caption{PePE Reconstruction}
\label{alg:pepe_reconstruction}
\KwIn{SfM points $\boldsymbol{\mu}$, Set of multiview images $I$, Corresponding set of masks $\mathcal{M}$ and monodepth maps $D$}
\KwOut{Optimized 3D Gaussians guided by Perception Prior}
\BlankLine

\textbf{[Step 1] Initialize Covariances, Colors, and Opacities.}\;
$(\mathbf{S}, \mathbf{c}, \boldsymbol{\sigma}) \leftarrow \text{InitAttributes()}$\;
$i \leftarrow 0$ \tcp*{Iteration Count}

\While{not converged}{
    \textbf{[Step 2] Set camera pose and image.}\;
    $(V, I) \leftarrow \text{SampleTrainingView()}$\;
    $\hat{I} \leftarrow \text{Rasterize}(\boldsymbol{\mu}, \mathbf{S}, \mathbf{c}, \boldsymbol{\sigma}, V)$\;

    \textbf{[Step 3] \underline{Mono-Depth Constrained Learning.}}\;
    $\boldsymbol{\mu}^{\mathrm{view}} \leftarrow \text{Transform}_{world \rightarrow view}(\boldsymbol{\mu})$\;
    $\mathbf{z} \leftarrow (\boldsymbol{\mu}^{\mathrm{view}})_z$\;
    $\hat{\mathcal{D}} \leftarrow \text{Rasterize}(\boldsymbol{\mu}, \mathbf{S}, \mathbf{z}, \boldsymbol{\sigma}, V)$ \tcp*{alpha-blending on z-coordinate}
    $L \leftarrow \text{Loss}(I, \hat{I}, D, \hat{D})$\;
    $(\boldsymbol{\mu}, \mathbf{S}, \mathbf{c}, \boldsymbol{\sigma}) \leftarrow \text{Adam}(\nabla L)$\;

    \textbf{[Step 4] Density control.}\;
    \If{IsDensityControlIteration$(i)$}{
        \ForAll{Gaussians with $(\boldsymbol{\mu}_i, \boldsymbol{\Sigma}_i, \mathbf{c}_i, \boldsymbol{\sigma}_i)$}{
            Do Pruning \& Densification\;
        }
    }

    \textbf{[Step 5] \underline{Gaussian Refinement Guided Initialization.}}\;
    \If{IsGaussianRefinementIteration$(i)$}{
        $M \leftarrow \text{RandomSample}(\mathcal{M})$\;
        \ForAll{Gaussian $g_i$ with $\boldsymbol{\mu}_i$ projected to pixel $p$}{
            $e_i \leftarrow \left| \hat{D}(p) - \mathbf{z_i} \right|$\;
        }
        $\mathcal{G}_r \leftarrow \text{TopK}_i(-e_i)$\;
        
        \ForAll{Gaussian $g_i \in \mathcal{G}_r$}{
            $g_i' \leftarrow \text{Project}(g_i)$\;
            \If{$g_i'\ \text{overlapped on boundary of} \ M$}{
                $g_i^* \leftarrow \text{Refine}(g_i, M)$\;
                $g_i \leftarrow g_i^*$ \tcp*{Boundary Aware Refinement}
            }
        }
    }
    $i \leftarrow i + 1$\;
}
\end{algorithm}

\section{Detailed algorithm of PePE Reconstruction}
In \Cref{alg:pepe_reconstruction}, we present a detailed pseudo-algorithm that integrates the original 3D Gaussian Splatting (3DGS) framework with the proposed PePE Reconstruction. Specifically, we incorporate two key components: Gaussian refinement guided initialization (Step 5, Lines 18–28) and monocular-depth constrained learning (Step 3, Lines 8–13).
By integrating the Gaussian refinement step into the training pipeline, our method achieves better boundary alignment from the early stages of optimization, which is clearly distinct from prior approaches. In addition, the mono-depth constrained learning encourages the Gaussians to align with the scene geometry by supervising their alpha-rendered depth.
Note that the order of operations in the training loop is critical. The Gaussian refinement is applied after the gradient computation (Line 13 in Step 3) and after density control (Step 4). This scheduling prevents incorrect gradient propagation through the newly refined Gaussians, which could otherwise degrade the training stability and overall performance.

\section{Per-Scale Analysis}

To characterize how the feature field behaves as a function of the query scale itself, we additionally evaluate LERF-Mask-Fine at a set of fixed query scales in \Cref{tab:appendix_perscale}.
Fine-part accuracy peaks at $s = 0.01$ and decreases monotonically as the query scale grows.
The clustering statistics explain this behavior.
As $s$ increases, the number of HDBSCAN clusters drops from 83 to 43 while the average mask size grows from 2,336 to 5,058 pixels, showing that fine parts are progressively merged into the objects that contain them.
This is the intended behavior of a continuous scale formulation, and it confirms that the learned scale gate produces a genuine granularity hierarchy.

\begin{table}[h]
    \begin{center}    
    \resizebox{0.7\linewidth}{!}{%
    \begin{tabular}{l ccccc}
        \toprule
        \textbf{Query scale} $s$ & 0.01 & 0.3 & 0.5 & 0.7 & 1.0 \\
        \midrule
        Fine-part mIoU (\%) & \textbf{68.4} & 65.8 & 48.6 & 27.1 & 19.2 \\
        \midrule
        \# clusters & 83 & 79 & 61 & 49 & 43 \\
        Mean mask size (px) & 2,336 & 2,337 & 3,439 & 3,910 & 5,058 \\
        \bottomrule
    \end{tabular}}
    \end{center}
    \caption{Per-scale analysis of PePESeg3D on LERF-Mask-Fine under a fixed query scale. Top: fine-part mIoU. Bottom: granularity of the HDBSCAN clusters extracted from the rendered features. 
    }
    \label{tab:appendix_perscale}
\end{table}

\section{Efficiency Analysis}

We compare training time, peak memory, and inference latency across methods in \Cref{tab:appendix_cost}, and further decompose the training time of our pipeline into individual modules in \Cref{tab:appendix_breakdown}.
All measurements are conducted on SPIn-NeRF using a single NVIDIA RTX 4090.
Single-scale methods optimize geometry and segmentation jointly, so we report a single training time for them.
Multi-scale methods, including ours, train the two stages sequentially and are reported separately.
In \Cref{tab:appendix_breakdown}, \textit{other} denotes the cost shared with the 3DGS baseline, covering rasterization, the photometric loss, and back-propagation.

Comparing the \textbf{Recon} column in \Cref{tab:appendix_cost}, reconstruction adds under four minutes over SAGA.
\Cref{tab:appendix_breakdown} shows that the refinement and the depth loss account for 1.54 minutes of this, while the remainder comes from rendering and back-propagating the additional depth channel.
In segmentation, the \textbf{Seg} column in \Cref{tab:appendix_cost} shows an increase of 5.5 minutes over SAGA.
Most of the extra cost comes from the perception loss, which also drives the largest gain in \Cref{tab:exp_ablation_segmentation}.
Overall, PePESeg3D requires only 1 GiB more VRAM and modest additional training than SAGA, while maintaining comparable inference speed.

\begin{table}[t]
    \begin{center}  
    \begin{minipage}[b]{0.66\linewidth}
        \centering
        \resizebox{\linewidth}{!}{%
        \begin{tabular}{l cccc}
            \toprule
            \textbf{Method} & \textbf{VRAM} (GiB) & \textbf{Recon} (min) & \textbf{Seg} (min) & \textbf{Infer} (ms) \\
            \midrule
            Gaussian Grouping~\cite{gaussian_grouping} & 20.3 & \multicolumn{2}{c}{32.3} & 7.0 \\
            Unified-Lift~\cite{zhu2025rethinking} & 7.0 & \multicolumn{2}{c}{71.9} & 5.2 \\
            \midrule
            OmniSeg3D-GS~\cite{ying2024omniseg3d} & 4.0 & 14.9 & 29.5 & 2.9 \\
            SAGA~\cite{cen2025segment} & 12.1 & 7.7 & 15.4 & 6.8 \\
            \textbf{PePESeg3D (Ours)} & 13.1 & 11.2 & 20.9 & 6.6 \\
            \bottomrule
        \end{tabular}}
        \vspace{2pt}
        \caption{Cross-method cost on SPIn-NeRF. Single-scale methods train reconstruction and segmentation jointly.}
        \label{tab:appendix_cost}
    \end{minipage}
    \hfill
    \begin{minipage}[b]{0.3\linewidth}
        \centering
        \resizebox{\linewidth}{!}{%
        \begin{tabular}{l c}
            \toprule
            \textbf{Component} & \textbf{Time} (min) \\
            \midrule
            Recon: \textit{other} & 9.71 \\
            Recon: refinement & 0.27 \\
            Recon: depth & 1.27 \\
            \midrule
            Seg: \textit{other} & 15.80 \\
            Seg: scale-aware & 0.62 \\
            Seg: perception & 4.04 \\
            Seg: centroid & 0.34 \\
            \bottomrule
        \end{tabular}}
        \vspace{2pt}
        \caption{Per-module breakdown of PePESeg3D.}
        \label{tab:appendix_breakdown}
    \end{minipage}
    \end{center}  
\end{table}

\section{Robustness to 2D Priors and Limited Views}

Since PePESeg3D builds on masks and monocular depth obtained from pretrained 2D models, we examine how sensitive the framework is to the choice of these priors.
In \Cref{tab:appendix_robustness}, we replace SAM with SAM2~\cite{sam2}, which reduces mask supervision by generating 25--30\% fewer masks.
PePESeg3D remains more robust than SAGA under this weaker supervision, with an even larger performance gap.
In \Cref{tab:appendix_hyperparam}, we replace Depth-Anything-V2 with Depth-Anything-V3~\cite{da3} and MiDaS~\cite{midas}.
Performance varies by at most 0.8 mIoU and 0.9 mBIoU, demonstrating that PePESeg3D is also stable across depth backbones.
Depth-Anything-V2 nonetheless performs best in our setting, which we attribute to its sharper preservation of intricate details within each view.
We also simulate more challenging settings in \Cref{tab:appendix_robustness} by training with only $1/4$ of the input views.
PePESeg3D degrades less than SAGA, since the perception priors supply supervision at both stages.

\section{Hyperparameter Sensitivity}

\Cref{tab:appendix_hyperparam} varies the refinement ratio $k$, the centroid similarity threshold $\tau$, the centroid refresh period $T$, and the depth backbone on LERF-Mask.
All variants stay within 0.5\%p of the default, confirming robustness to hyperparameter choices.
The refinement ratio $k$ controls how many Gaussians are considered as surface candidates.
Setting $k = 5$ leaves boundary Gaussians unrefined, while $k = 20$ admits off-surface Gaussians whose refinement perturbs the geometry, and both cause a minor drop.
Changing $\tau$ or $T$ shifts the results by under 0.5\%p, indicating that the centroid loss is stable with respect to the matching threshold and the refresh interval.

\begin{table}[t]
    \begin{center}
    \resizebox{0.6\linewidth}{!}{%
    \begin{tabular}{l ccc}
        \toprule
        \textbf{Method} & \textbf{Default} & \textbf{1/4 views} & \textbf{SAM2}~\cite{sam2} \\
        \midrule
        SAGA~\cite{cen2025segment} & 69.6 / 67.8 & 60.5 / 58.5 & 64.6 / 62.5 \\
        \textbf{PePESeg3D (Ours)} & \textbf{70.8 / 68.3} & \textbf{62.9 / 59.6} & \textbf{67.1 / 65.5} \\
        \midrule
        Gap & +1.2 / +0.5 & +2.4 / +1.1 & +2.5 / +3.0 \\
        \bottomrule
    \end{tabular}}
    \end{center}
    \caption{Robustness to weaker 2D supervision and limited view coverage on LERF-Mask-Fine, reported as mIoU / mBIoU (\%).}
    \label{tab:appendix_robustness}
\end{table}

\begin{table}[t]
    \begin{center}
    \resizebox{0.9\linewidth}{!}{%
    \begin{tabular}{l c cc cc cc cc}
        \toprule
        & \textbf{Default} & \multicolumn{2}{c}{Refinement $k$} & \multicolumn{2}{c}{Centroid $\tau$} & \multicolumn{2}{c}{Refresh $T$} & \multicolumn{2}{c}{Depth backbone} \\
        \cmidrule(lr){3-4}\cmidrule(lr){5-6}\cmidrule(lr){7-8}\cmidrule(lr){9-10}
        & & $5$ & $20$ & $0.85$ & $0.95$ & $100$ & $400$ & DA-V3 & MiDaS \\
        \midrule
        mIoU  & \textbf{80.5} & 80.3 & 80.2 & 80.5 & 80.5 & 80.4 & 80.4 & 79.8 & 79.7 \\
        mBIoU & \textbf{76.5} & 76.3 & 76.0 & 76.4 & 76.5 & 76.1 & 76.3 & 75.7 & 75.6 \\
        \bottomrule
    \end{tabular}}    
    \end{center}
    \caption{Sensitivity to the refinement ratio $k$, the centroid threshold $\tau$, the refresh period $T$, and the depth backbone on LERF-Mask. The default configuration is $k = 10\%$, $\tau = 0.90$, $T = 200$, and Depth-Anything-V2.}
    \label{tab:appendix_hyperparam}
\end{table}

\begin{figure*}[t!]
    \centering
    \includegraphics[width=\linewidth]{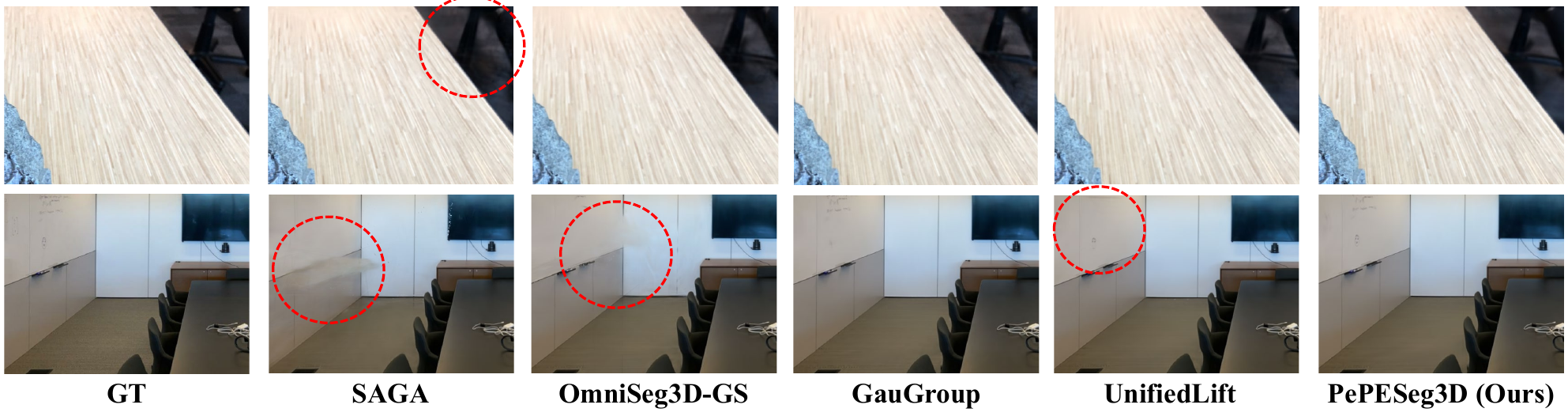}
    \caption{Additional comparison of reconstruction quality on SPIn-NeRF. The red circles indicate artifacts and missing details present in the baseline methods.}
    \label{fig:appendix_pepe_recon_additional}
    \vspace{-1.0em}
\end{figure*}

\begin{figure}[h!]
  \centering
    \includegraphics[width=0.6\linewidth]{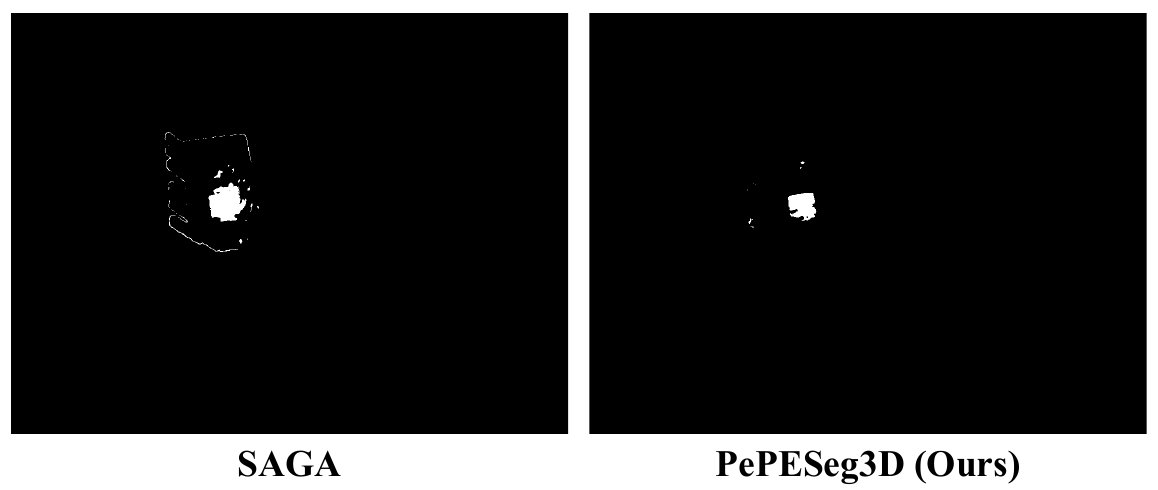}
  \caption{Uncropped segmentation result corresponding to \Cref{fig:qualitative_vertical}-(b). The red dot in the lower-bound scale of \Cref{fig:qualitative_vertical}-(b) indicates the point prompt. The results presented in the main paper show a zoomed-in view of the green boxed region in the lower-bound scale.}
  \label{fig:appendix_uncropped_finegrained}
  \vspace{-1.0em}
\end{figure}

\section{More Qualitative Results}
\begin{sidewaysfigure*}[p] 
    \centering
    \includegraphics[width=\textwidth]{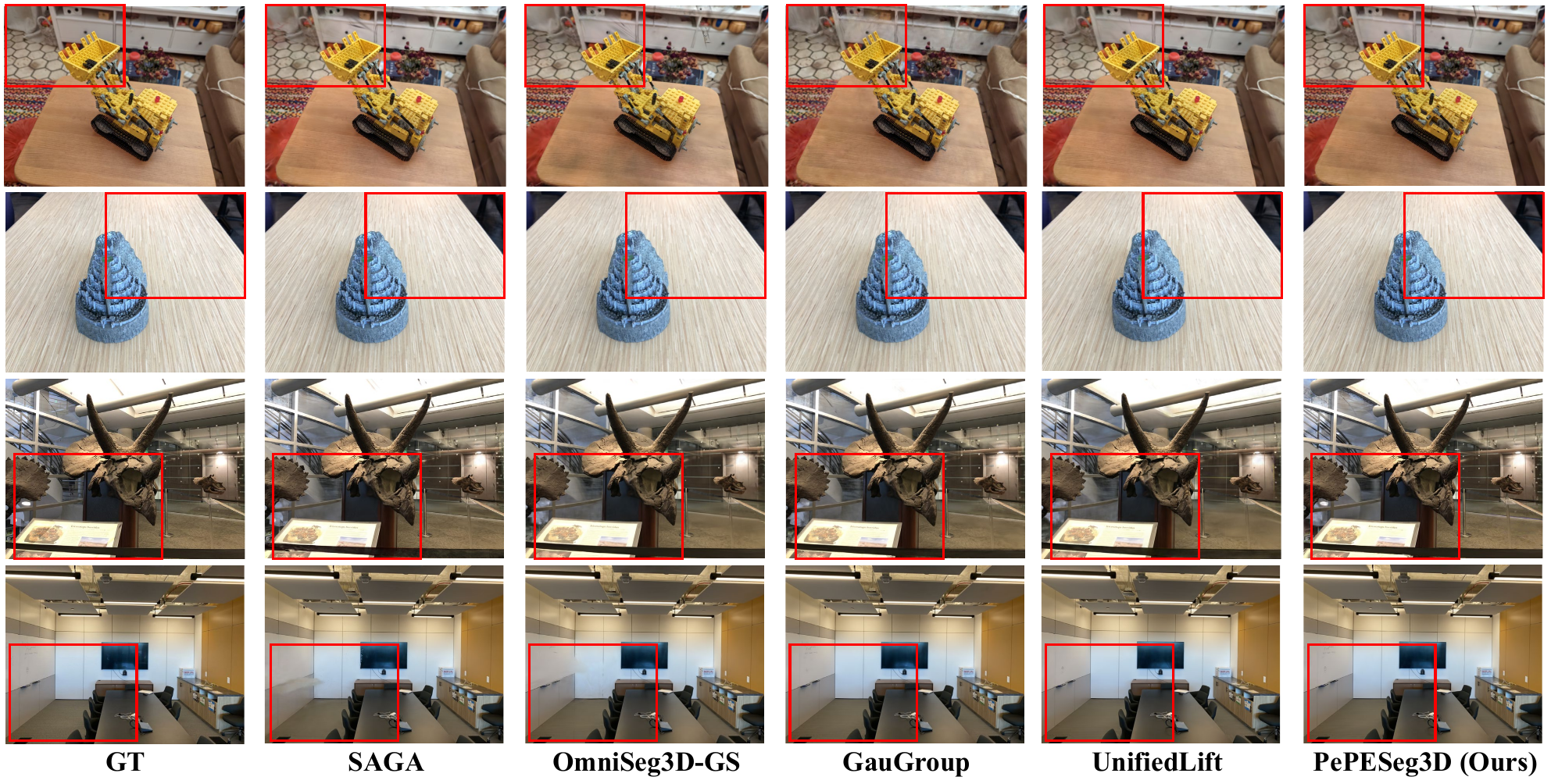}
    \vspace{0.5em}
    \caption{Uncropped reconstruction results corresponding to \Cref{fig:exp_spin_nerf_joint,fig:appendix_pepe_recon_additional}. The red boxes indicate the cropped areas presented in each corresponding figure.}
    \label{fig:appendix_pepe_recon_uncropped}
\end{sidewaysfigure*}

\begin{figure*}[t!]
    \centering
    \includegraphics[width=0.90\linewidth]{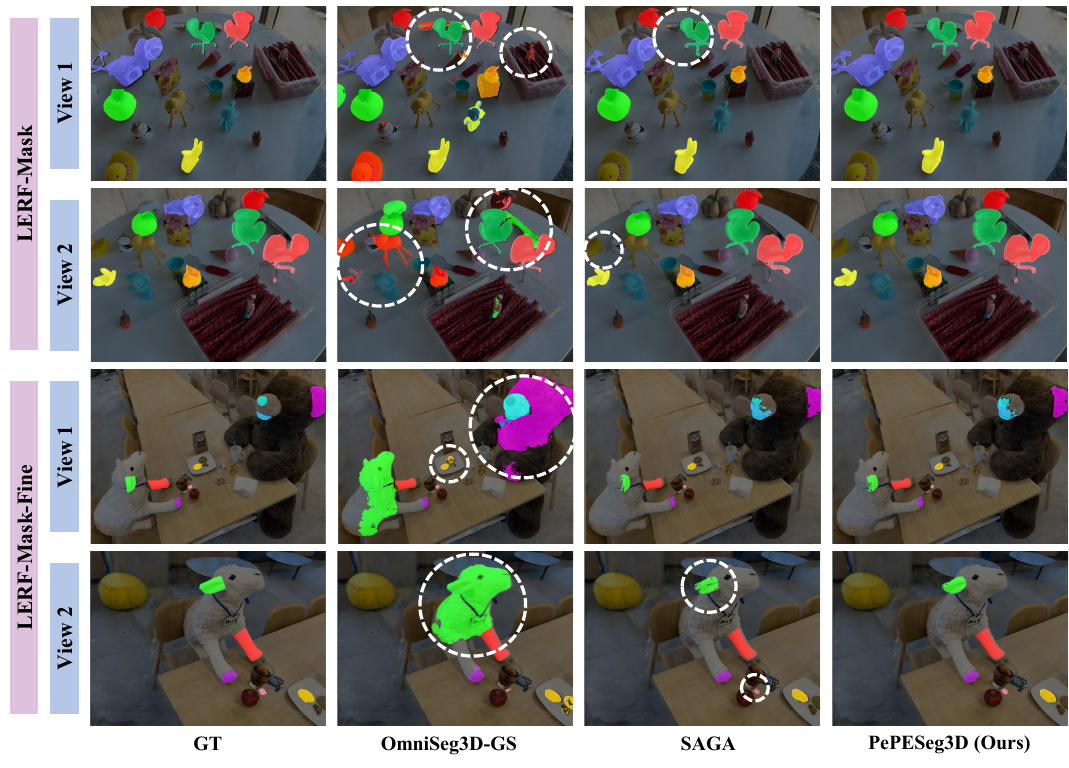}
    \caption{Visual comparison of segmentation results on the LERF-Mask and LERF-Mask-Fine datasets. OmniSeg3D-GS struggles with view consistency and tends to assign identical labels to semantically distinct objects. While SAGA shows reasonable performance on LERF-Mask despite minor over-segmentation, it often overlooks fine-grained details, such as the small tag in LERF-Mask-Fine. PePESeg3D effectively addresses these limitations, producing consistent and accurate segmentations.}
    \label{fig:appendix_lerf_mask_segmentation}
\end{figure*}

We provide additional qualitative comparisons to further validate the effectiveness of PePESeg3D. 
Extended reconstruction comparisons with baseline methods are presented in \Cref{fig:appendix_pepe_recon_additional}.
To provide full context and improve clarity, \Cref{fig:appendix_pepe_recon_uncropped,fig:appendix_uncropped_finegrained} present uncropped, high-resolution versions of the visual results discussed in the main paper and \Cref{fig:appendix_pepe_recon_additional}. 
In the main text, only the most relevant regions were highlighted and reported to support key discussion points.
In \Cref{fig:appendix_lerf_mask_segmentation}, we show segmentation results from various viewpoints on the LERF-Mask and LERF-Mask-Fine datasets. These results further illustrate the superior view consistency and mask quality achieved by our method compared to the baselines.

\begin{figure*}[t]
  \centering
    \includegraphics[width=0.8\linewidth]{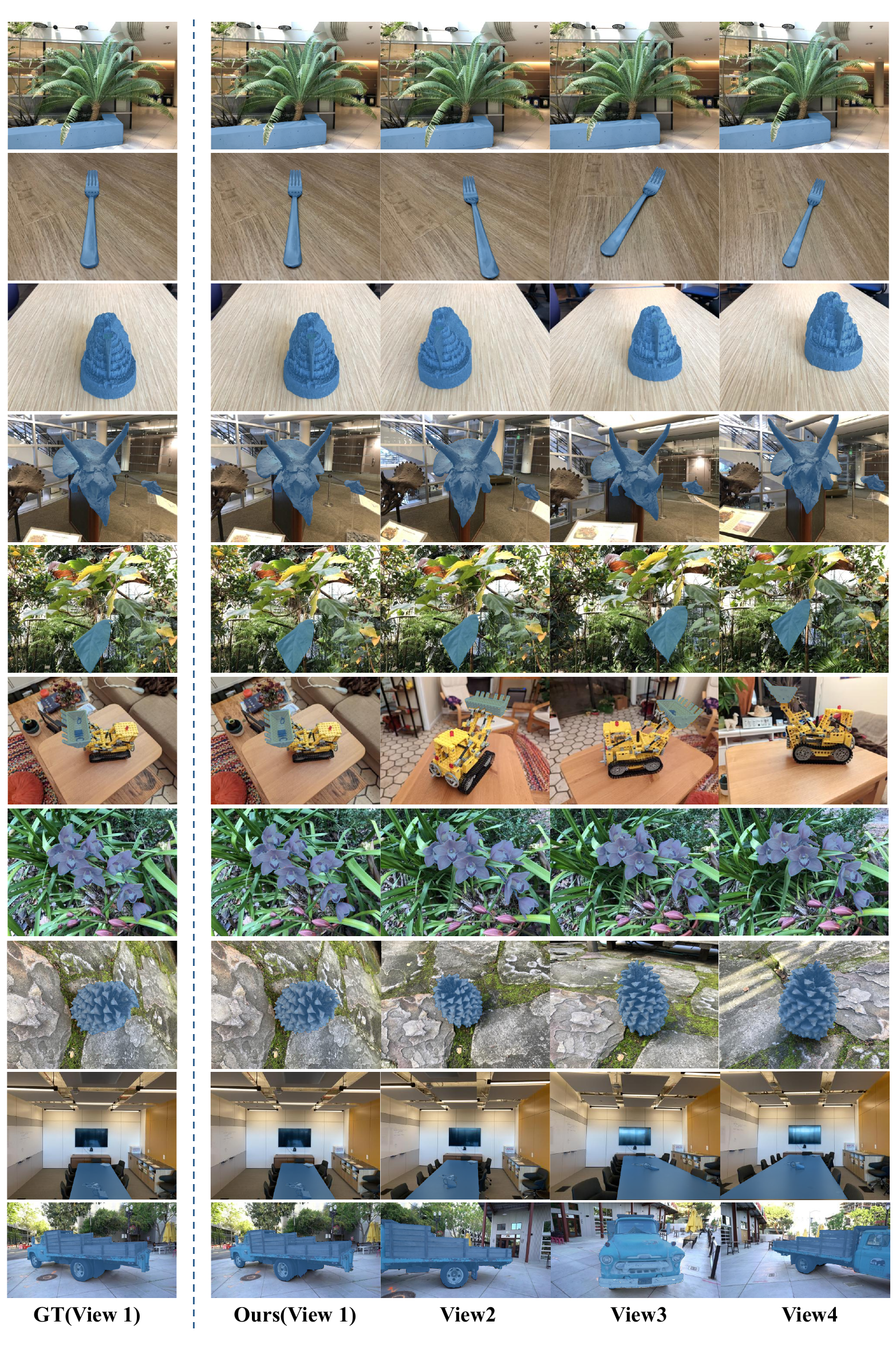}
  \caption{SPIn-NeRF segmentation results given in multi-view generated by PePESeg3D.}
  \label{fig:appendix_onlypepe_spin_nerf_multiview}
\end{figure*}

\begin{figure*}[t]
    \centering
    \includegraphics[width=0.9\linewidth]{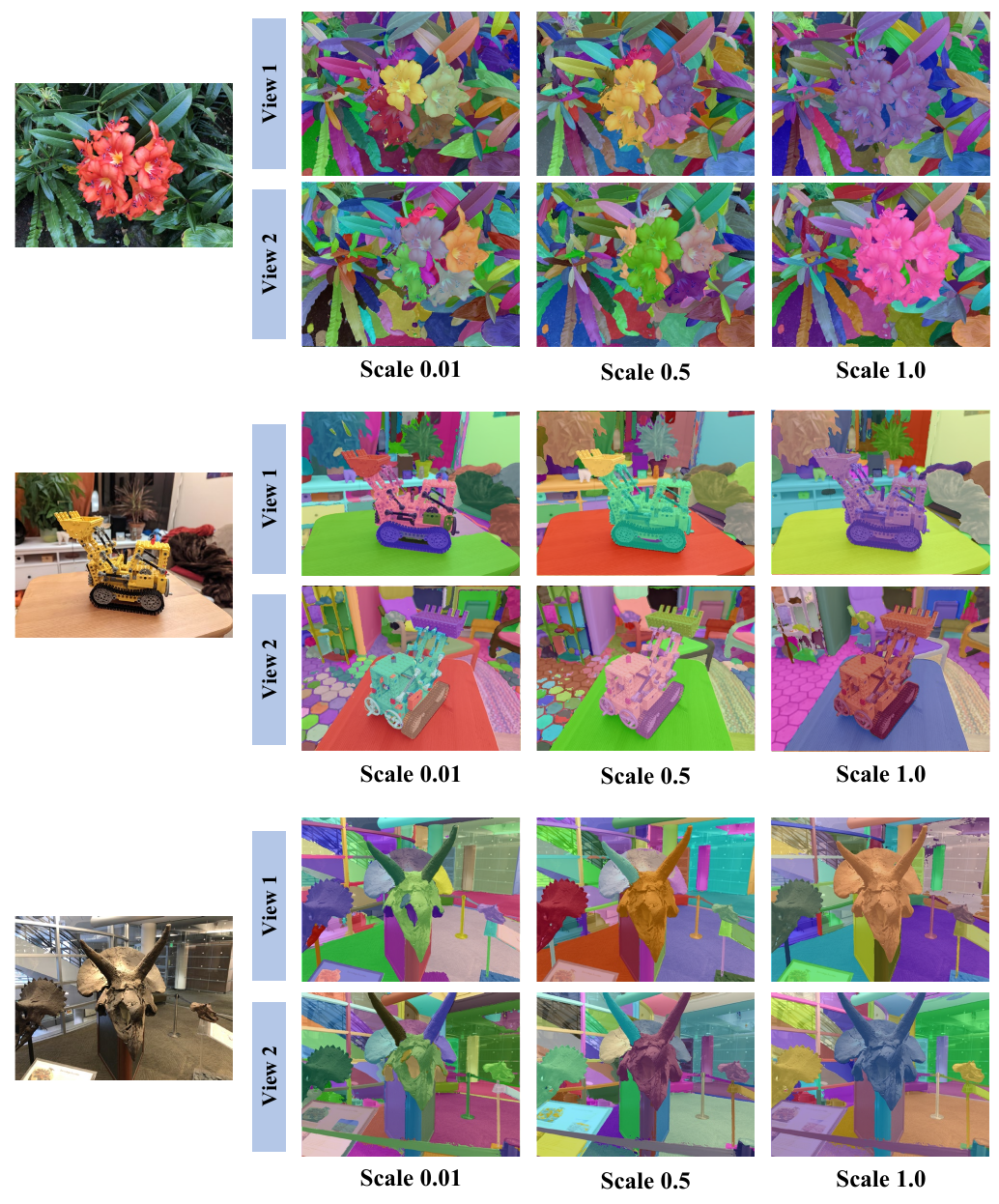}
    \caption{Automatic Decomposition Visualization on SPIn-NeRF generated by PePESeg3D. Random colors are assigned to semantic clusters derived via HDBSCAN.
    The results demonstrate appropriate scale-dependent granularity while maintaining robust view consistency across different perspectives.}
    \label{fig:appendix_automatic}
\end{figure*}

Furthermore, \Cref{fig:appendix_onlypepe_spin_nerf_multiview} visualizes the segmentation results for all scenes in the SPIn-NeRF dataset. 
The first column displays the ground truth masks, followed by our model's predictions in the second column. 
The subsequent three columns showcase segmentation results from alternate viewpoints, highlighting the robustness and consistency of our feature field across different perspectives.
Finally, \Cref{fig:appendix_automatic} presents automatic scene decomposition results across three distinct scale levels.
For the automatic scene decomposition task, we apply HDBSCAN clustering to the 2D rendered features and assign a random color to each label.

\begin{figure}[t!]
    \centering
    \includegraphics[width=0.6\linewidth]{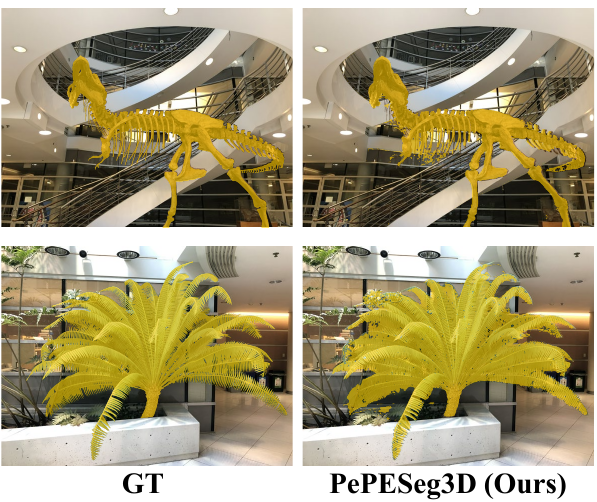}
    \caption{Failure Cases in NVOS}
    \label{fig:appendix_failure_case}
\end{figure}

\section{Limitation}
\paragraph{Scale Coupling Effects}
We observe that the scale gate mechanism leads to cross-scale interactions, where features selected at a certain scale can affect segmentation representations at other scales.
When upper-scale features are selected, the associated channels may also be activated at lower scales because the gating mechanism operates on a common feature space across different scales.
This leads to a situation where masks learned at lower scales influence upper-scale predictions, and vice versa.
While such cross-scale feature sharing can enrich the representation, features rendered at nearby query scales become similar to each other.
As a result, a given scale may not cleanly separate two entities whose physical scales are almost identical.
A promising direction for future work is to develop more effective mechanisms for isolating or controlling these interactions.

\paragraph{Segmentation Incompleteness in Complicated Details}
Although we employ perceptual loss to mitigate the inherent incompleteness of SAM masks, segmentation errors still occur when encountering objects with highly complicated structures, as shown in~\Cref{fig:appendix_failure_case}.
For intricate targets that SAM struggles to recognize as a single entity, the model may occasionally fail to achieve precise segmentation.
This suggests that while perceptual priors help, they cannot fully compensate for the underlying backbone's limitation in zero-shot segmentation.
To address these limitations, future efforts could focus on enhancing the model's ability to unify fragmented predictions by incorporating modules that specifically reason about the global connectivity of a mask.

\paragraph{Scene-Dependent Scale Normalization}
Since the physical scale is normalized within each scene, the same query scale corresponds to different absolute granularities across scenes, and the boundaries between granularity levels shift accordingly.
Selecting a query scale for a new scene thus requires either a reference object or a short sweep, and learning a scene-agnostic parameterization of scale is left for future work.

\end{document}